\documentclass[letterpaper, 10 pt, conference]{ieeeconf}  

\IEEEoverridecommandlockouts                              

\usepackage{graphics} 
\usepackage{epsfig} 
\usepackage{mathptmx} 
\usepackage{times} 
\usepackage{amsmath} 
\usepackage{amssymb}  
\usepackage[ruled]{algorithm2e}
\usepackage{subfigure}
\usepackage{tabu}
\usepackage{booktabs}
\usepackage{multirow}
\usepackage{footnote}
\usepackage{threeparttable}
\usepackage{booktabs}
\usepackage[table]{xcolor}
\usepackage[letterpaper,top=60pt,bottom=43pt,left=48pt,right=48pt]{geometry}

\title{\LARGE \bf
Lifelong Federated Reinforcement Learning: A Learning Architecture for Navigation in Cloud Robotic Systems
}

\author{Boyi Liu$^{{1},{3}}$, Lujia Wang$^{1}$ and Ming Liu$^{2}$
\thanks{*This work was supported by National Natural Science Foundation of China No. 61603376; Guangdong-Hongkong Joint Scheme (Y86400); Shenzhen Science, Technology and Innovation Commission (SZSTI) Y79804101S awarded to Dr. Lujia Wang.}
\thanks{$^{1}$Boyi liu, Lujia Wang are with Cloud Computing Lab of Shenzhen Institutes of Advanced Technology, Chinese Academy of Sciences.{\tt\small liuboyi17@mails.ucas.edu.cn};
	{\tt\small lj.wang1@siat.ac.cn};
	{\tt\small cz.xu@siat.ac.cn}}
\thanks{$^{2}$Ming liu, is with Department of ECE, Hong Kong University of Science and Technology. {\tt\small eelium@ust.hk}}
\thanks{$^{3}$Boyi liu is also with the University of Chinese Academy of Sciences.}
}

\begin{document}

\maketitle
\thispagestyle{empty}
\pagestyle{empty}

\begin{abstract}
This paper was motivated by the problem of how to make robots fuse and transfer their experience so that they can effectively use prior knowledge and quickly adapt to new environments. To address the problem, we present a learning architecture for navigation in cloud robotic systems: Lifelong Federated Reinforcement Learning (LFRL). In the work, we propose a knowledge fusion algorithm for upgrading a shared model deployed on the cloud. Then, effective transfer learning methods in LFRL are introduced. LFRL is consistent with human cognitive science and fits well in cloud robotic systems. Experiments show that LFRL greatly improves the efficiency of reinforcement learning for robot navigation. The cloud robotic system deployment also shows that LFRL is capable of fusing prior knowledge. In addition, we release a cloud robotic navigation-learning website to provide the service based on LFRL: \emph{www.shared-robotics.com}.
\end{abstract}

\section{INTRODUCTION}
Autonomous navigation is one of the core issues in mobile robotics. It is raised among various techniques of avoiding obstacles and reaching target position for mobile robotic navigation. Recently, reinforcement learning (RL) algorithms are widely used to tackle the task of navigation. RL is a kind of reactive navigation method, which is an important meaning to improve the real-time performance and adaptability of mobile robots in unknown environments. Nevertheless, there still exists a number of problems in the application of reinforcement learning in navigation such as reducing training time, storing data over long time, separating from computation, adapting rapidly to new environments etc [1].

\textbf{In this paper, we address the problem of how to make robots learn efficiently in a new environment and extend their experience so that they can effectively use prior knowledge.} We focus on cloud computing and cloud robotic technologies [2], which can enhance robotic systems by facilitating the process of sharing trajectories, control policies and outcomes of collective robot learning. Inspired by human congnitive science present in Fig.1, we propose a \textbf{L}ifelong \textbf{F}ederated \textbf{R}einforcement \textbf{L}earning (LFRL) architecture to realize the goal.
\begin{figure}[thpb]
	\centering
	\includegraphics[width=0.8\linewidth]{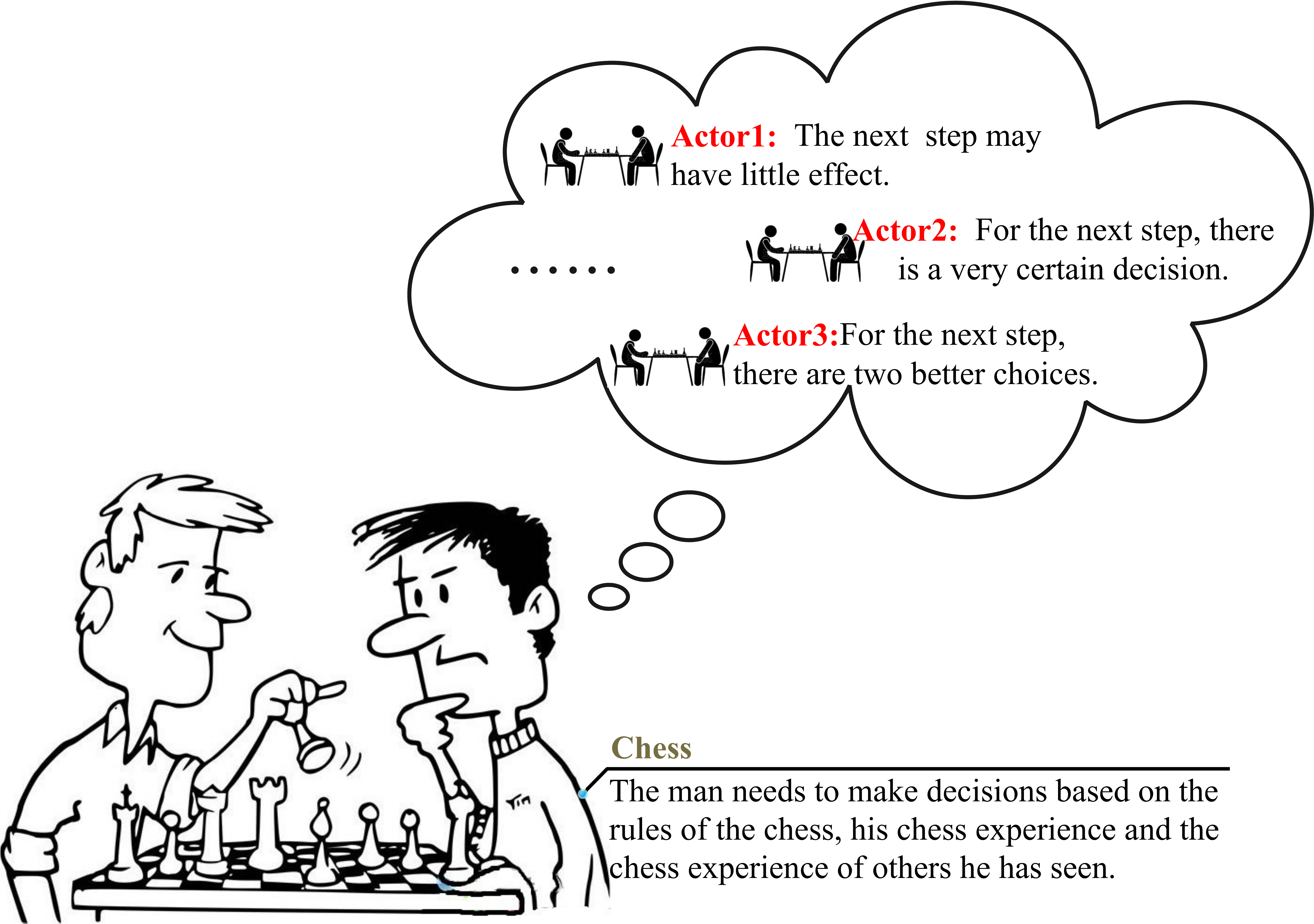}
	\caption{The person on the right is considering where should the next step go. The chess he has played and the chess he has seen are the most two influential factors on making decisions. His memory fused into his policy model. So how can robots remember and make decisions like humans? Motivated by this human cognitive science, we propose the LFRL in cloud robot systems. LFRL makes the cloud remember what robots learned before like a human brain.}
	\label{fig:architecture}
\end{figure}
With the scalable architecture and knowledge fusion algorithm, LFRL achieves exceptionally efficiency in reinforcement learning for cloud robot navigation. LFRL enables robots to remember what they have learned and what other robots have learned with cloud robotic systems. LFRL contains both asynchronization and synchronization learning rather than limited to synchronous learning as A3C [3] or UNREAL [4]. To demonstrate the efficacy of LFRL, we test LFRL in some public and self-made training environments. Experimental result indicates that LFRL is capable of enabling robots effectively use prior knowledge and quickly adapt to new environments. Overall, this paper makes the following contributions:
\begin{itemize}
\item We present a Lifelong Federated Reinforcement Learning architecture based on human cognitive science. It makes robots perform lifelong learning of navigation in cloud robotic systems. 
\item We propose a knowledge fusion algorithm. It is able to fuse prior knowledge of robots and evolve the shared model in cloud robotic systems. 
\item Two effective transfer learning approaches are introduced to make robots quickly adapt to new environments. 
\item A cloud robotic navigation-learning website is built in the work: \emph{www.shared-robotics.com}. It provides the service based on LFRL.
\end{itemize}
\section{Related Theory}
\subsection{Reinforcement learning for navigation}
Eliminating the requirements for location, mapping or path planning procedures, several DRL works have been presented that successful learning navigation policies can be achieved directly from raw sensor inputs: target-driven navigation [5], successor feature RL for transferring navigation policies [6], and using auxiliary tasks to boost DRL training [7]. Many follow-up works have also been proposed, such as embedding SLAM-like structure into DRL networks [8], or utilizing DRL for multi-robot collision avoidance [9]. Tai et al [10] successfully appplied DRL for mapless navigation by taking the sqarse 10-dimensional range findings and the target position , defining mobile robot coordinate frame as input and continuous steering commands as output. Zhu et al. [5] input both the first-person view and the image of the target object to the A3C model, formulating a target-driven navigation problem based on the universal value function approximators [11]. To make the robot learn to navigate, we adopt a reinforcement learning perspective, which is built on recent success of deep RL algorithms for solving challenging control tasks [12-15]. Zhang [16] presented a solution that can quickly adapt to new situations (e.g., changing navigation goals and environments). Making the robot quickly adapt to new situations is not enough, we also need to consider how to make robots capable of memory and evolution, which is similar to the main purpose of lifelong learning.

\subsection{Lifelong machine learning}
Lifelong machine learning, or LML [17], considers system that can learn many tasks from one or more domains over its lifetime. The goal is to sequentially store learned knowledge and to selectively transfer that knowledge when a robot learns a new task, so as to develop more accurate hypotheses or policies. Robots are confronted with different obstacles in different environments, including static and dynamic ones, which are similar to the multi-task learning in lifelong learning. Although learning tasks are the same, including reaching goals and avoiding obstacles, their obstacle types are different. There are static obstacles, dynamic obstacles, as well as different ways of movement in dynamic obstacles. Therefore, it can be regarded as a low-level multitasking learning.

A lifelong learning should be able to efficiently retain knowledge. This is typically done by sharing a representation among tasks, using distillation or a latent basis [18]. The agent should also learn to selectively use its past knowledge to solve new tasks efficiently. Most works have focused on a special transfer mechanism, i.e., they suggested learning differentiable weights are from a shared representation to the new tasks [4, 19]. In contrast, Brunskill and Li [20] suggested a temporal transfer mechanism, which identifies an optimal set of skills in new tasks. Finally, the agent should have a systematic approach that allows it to efficiently retain the knowledge of multiple tasks as well as an efficient mechanism to transfer knowledge for solving new tasks. Chen [21] proposed a lifelong learning system that has the ability to reuse and transfer knowledge from one task to another while efficiently retaining the previously learned knowledge-base in Minecraft. Although this method has achieved good results in Mincraft, there is a lack of multi-agent cooperative learning model. Learning different tasks in a same scene is similar but different for robot navigation learning.
\subsection{Federated learning}
LFRL realizes federated learning of multi robots through knowledge fusion. Federated learning was first proposed in [22], which showed its effectiveness through experiments on various datasets. In federated learning systems, the raw data is collected and stored at multiple edge nodes, and a machine learning model is trained from the distributed data without sending the raw data from the nodes to a central place [23, 24]. Different from the traditional joint learning method where multiple edges are learning at the same time, LFRL adopts the method of \textit{first training then fusing} to reduce the dependence on the quality of communication[25, 26].
\subsection{Cloud robotic system}
LFRL fits well with cloud robotic system. Cloud robotic system usually relies on many other resources from a network to support its operation. Since the concept of the cloud robot was proposed by Dr. Kuffner of Carnegie Mellon University (now working at Google company) in 2010 [27], the research on cloud robots is rising gradually. At the beginning of 2011, the cloud robotic study program of RoboEarth [28] was initiated by the Eindhoven University of Technology. Google engineers have developed robot software based on the Android platform, which can be used for remote control based on the Lego mind-storms, iRobot Create and Vex Pro, etc. [29]. Wang et al. present a framework targeting near real-time MSDR, which grants asynchronous access to the cloud from the robots [30]. However, no specific navigation method for cloud robots has been proposed up to now. We believe that this is the first navigation learning architecture for cloud robotic systems. 

Generally, this paper focuses on developing a reinforcement learning architecture for robot navigation, which is capable of lifelong federated learning and multi robots federated learning. This architecture is well fit in cloud robot systems.
\section{Methodology}
\begin{figure*}[thpb]
	\centering
	\includegraphics[width=0.9\linewidth]{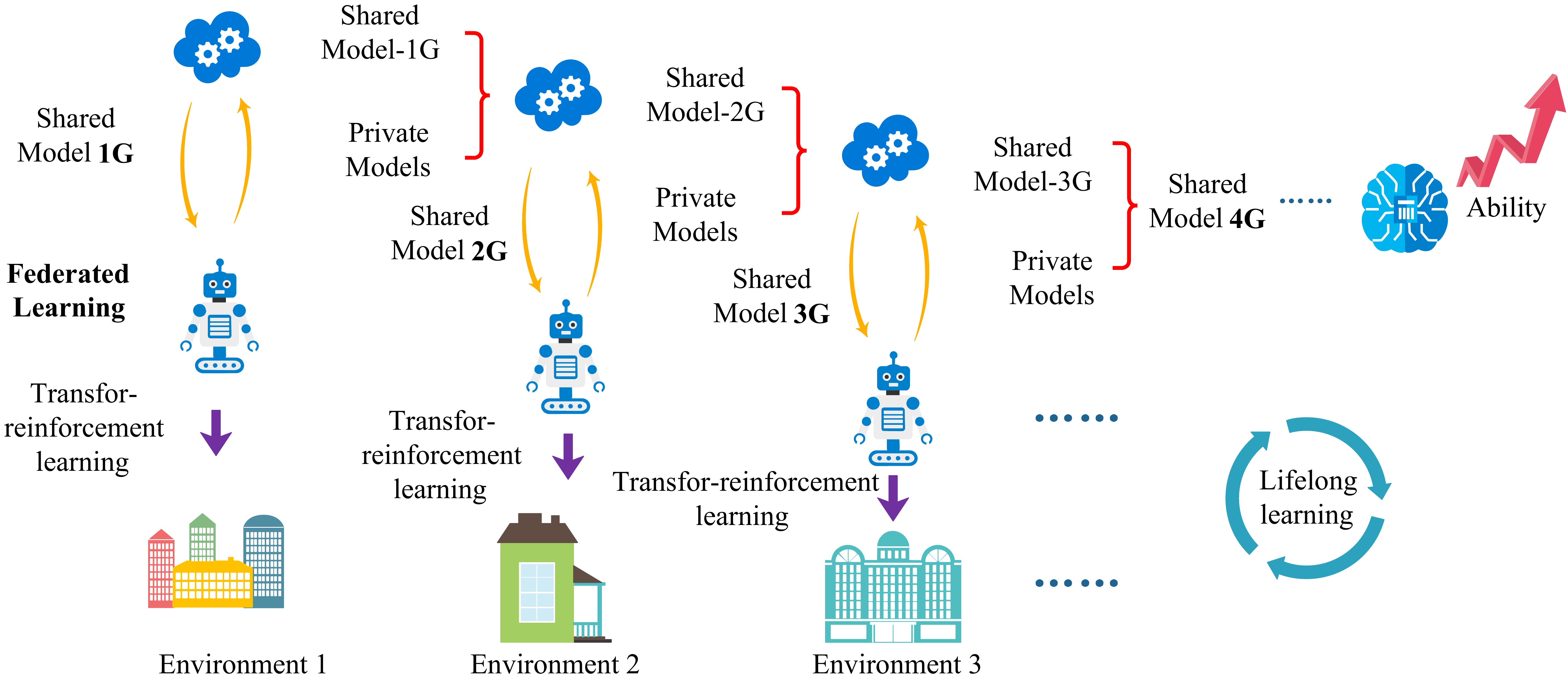}
	\caption{Proposed Architecture. In Robot$\rightarrow$Environment, the robot learns to avoid some new types of obstacles in the new environment through reinforcement learning and obtains the private Q-network model. Not only from one robot training in different environments, private models can also be resulted from multiple robots. It is a type of federated learning. After that, the private network will be uploaded to the cloud. The cloud server evolves the shared model by fusing private models to the shared model. In Cloud$\rightarrow$Robot, inspired by transfer learning, successor features are used to transfer the strategy to unknown environment. We input the output of the shared model as added features to the Q-network in reinforcement learning, or simply transfer all parameters to the Q-network. Iterating this step, models on the cloud become increasingly powerful.}
	\label{fig:architecture}
\end{figure*}
LFRL is capable of reducing training time without sacrificing accuracy of navigating decision in cloud robotic systems. LFRL uses Cloud-Robot-Environment setup to learn the navigation policy. LFRL consists of a cloud server, a set of environments, and one or more robots. We develop a federated learning algorithm to fuse private models into the shared model in the cloud. The cloud server fuses private models into the shared model, then evolves the shared model. As illustrated in Fig.2, LFRL is an implementation of lifelong learning for navigation in cloud robotic systems.
Compared with A3C or UNREAL approaches which update parameters of the policy network at the same time, the proposed knowledge fusion approach is more suitable for the federated architecture of LFRL. The proposed approach is capable of fusing models with asynchronous evolution. The approach of updating parameters at the same time has certain requirements for environments, while the proposed knowledge fusion algorithm has no requirements for environments. Using generative network and dynamic weight labels are able to realize the integration of memory instead of A3C or UNREAL method, which only generates a decision model during learning and has no memory.
\begin{figure}[thpb]
	\centering
	\includegraphics[width=0.5\linewidth]{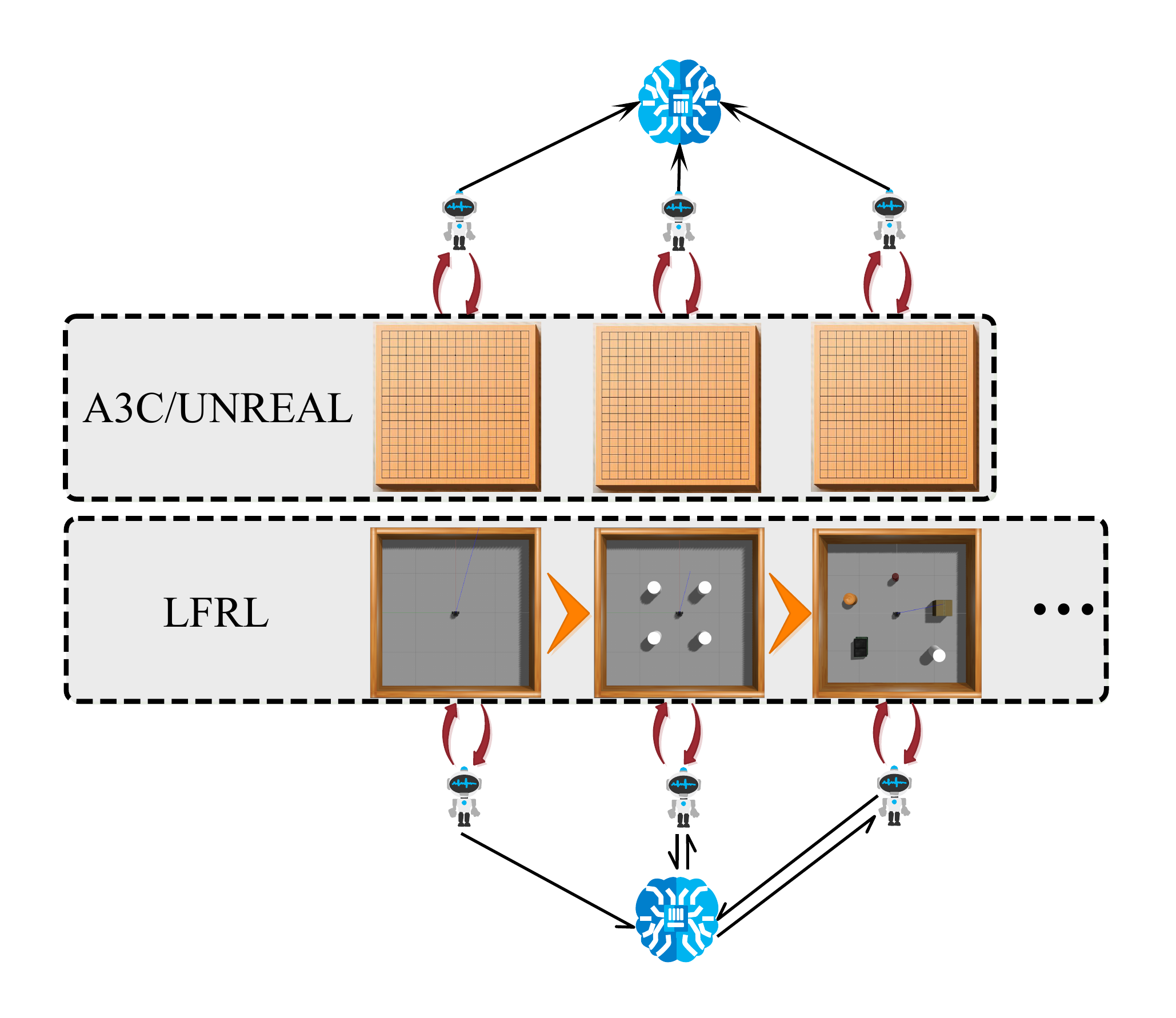}
	\caption{LFRL compared with A3C or UNREAL}
	\label{fig:architecture}
\end{figure}
As illustrated in Fig.3. In the algorithm of A3C or UNREAL, the training environment is constant. States of agents are countable. The central node only needs to fuse parameters, which can be performed at the same time. The two methods are capable of fusing parameters while training. Network structures of agents must be the same. However, in LFRL, the training environment is variable. State of agents are uncountable with more training environments uploading. In different agents, the structure of hiden layers of the policy network can be different. The cloud fuse training results. The robots are trained in new environments based on the shared model. Robots and the cloud have interactions in upload and download procedures. LFRL is more suitable for the cloud robot system where the environment is uncountable, especially in the lifelong learning framework.
\subsection{Procedure of LFRL}
This section displays a practical example of LFRL: there are 4 robots, 3 different environments and cloud servers. The first robot obtains its private strategy model Q1 through reinforcement learning in Environment 1 and upload it to the cloud server as the shared model 1G. After a while, Robot 2 and Robot 3 desire to learn navigation by reinforcement learning in Environment 2 and Environment 3. In LFRL, Robot 2 and Robot 3 download the shared model 1G as the initial actor model in reinforcement learning. Then they can get their private networks Q2 and Q3 through reinforcement learning in Environment 2 and Environment 3. After completing the training, LFRL uploads Q2 and Q3 to the cloud. In the cloud, strategy models Q2 and Q3 will be fused into shared model 1G, and then shared model 2G will be generated. In the future, the shared model 2G can be used by other cloud robots. Other robots will also upload their private strategy models to the cloud server to promote the evolution of the shared model.

The more complicated tasks responded to more kinds of obstacles in robot navigation. The learning environment is gigantic in robot navigation learning. This case is different from the chess. So we borrow the idea of lifelong learning. Local robots will learn to avoid more kinds of obstacles and the cloud will fuse these skills. These skills will be used in more defined and undefined environments. For a cloud robotic system, the cloud generates a shared model for a time, which means an evolution in lifelong learning. The continuous evolution of the shared model in cloud is a lifelong learning pattern. In LFRL, the cloud server achieves the knowledge storage and fusion of a robot in different environments. Thus, the shared model becomes powerful through fusing the skills to avoid multi types of obstacles.

For an individual robot, when the robot downloads the cloud model, the initial Q-network has been defined. Therefore, the initial Q-network has the ability to reach the target and avoid some types of obstacles. It is conceivable that LFRL can reduce the training time for robots to learn navigation. Furthermore, there is a surprising experiment result that the robot can get higher scores in navigation with LFRL.
\begin{algorithm}
	\caption{Processing Algorithm in LFRL}
	Initialize action-value Q-network with random weights ${\theta }$\;
	\KwIn{${\theta }_{a}$: The parameters of the a-th shared model in cloud ; $m$: The number of private networks.} 
	\KwOut{The evolved ${\theta }_{a}$}
	\While{cloud server is running}{
		\If{service\_request=True}
		{
			Transfer ${\theta }_{a}$ to $\pi$\;
			\For{$i=1;i \le m;i++$} 
			{ 
				
				${\theta }_{i}\leftarrow robot(i)$ perform reinforcement learning with $\pi$ in environment.\;
				
				Send ${\theta }_{i}$ to cloud\;
			} 
		}
		\If{evolve time=True}
		{
			Generate ${\theta }_{a+1}$ = \emph{fuse}(${\theta }_{1},{\theta }_{2},\cdots,{\theta }_{m},{\theta }_{a}$)
			${\theta }_{a}\leftarrow {\theta }_{a+1}$
		}
		
	}
\end{algorithm}
However, in actual operation, the cloud does not necessarily fuse models every time it receives a private network, but fuses at a fixed frequency rate. So, we present the processing flow of LFRL shown in Algorithm 1.
\begin{figure*}[thpb]
	\centering
	\includegraphics[width=0.9\linewidth]{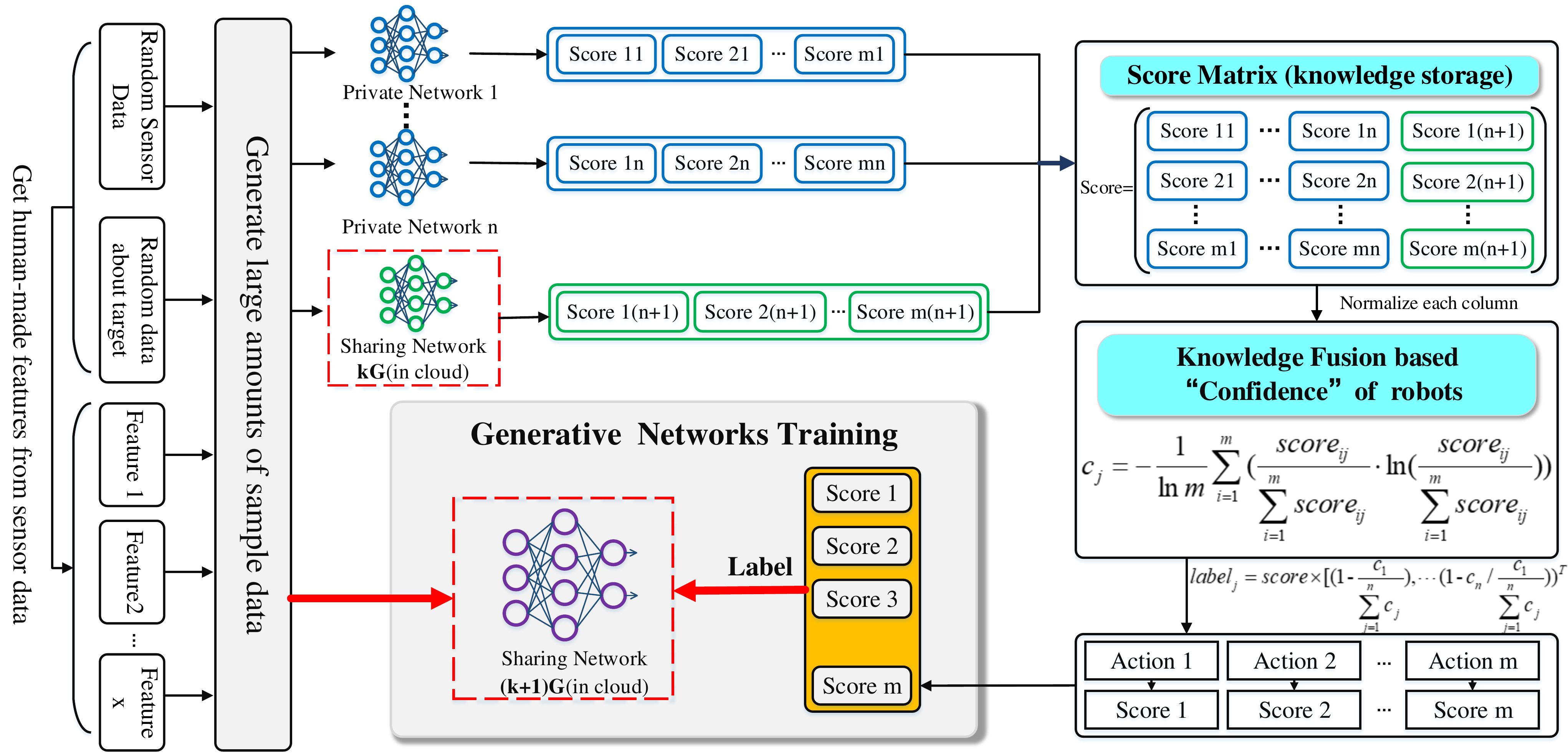}
	\caption{Knowledge Fusion Algorithm in LFRL: We generate a large amount of training data based on sensor data, target data, and human-defined features. Each training sample is added into the private network and the k-th generation sharing network, while different actors are scored for different actions. Then, we store the scores and calculate the confidence values of all actors in this training sample data. The ``confidence value'' is used as a weight, while the scores are weighted and summed to obtain the label of the current sample data. By analogy, all sample data labels are generated. Finally, a network is generated and fits the sample data as much as possible. The generated network is the (k+1)th generation. This step of fusion is finished.}
	\label{fig:architecture}
\end{figure*}
Key algorithms in LFRL include knowledge function algorithm and transferring approaches, as introduced in the following.
\subsection{Knowledge fusion algorithm in cloud}
Inspired by images style transfer algorithm, we develop a knowledge fusion algorithm to evolve the shared model. This algorithm is based on generative networks and it is efficient to fuse parameters of networks trained from different robots or a robot in different environments. The algorithm deployed in the cloud server receives the privately transmitted network and upgrades the sharing network parameters. To address knowledge fusion, the algorithm generates a new shared model from private models and the shared model in cloud. This new shared model is the evolved model.

Fig.4 illustrates the process of generating a policy network. The structure of the policy network has the same input and output dimensions with the private policy network. The number of outputs is equal to the number of action types that robots can act. The dimensions of input also correspond with sensor data and human-made features. The training data of the network is randomly generated based on sensor data attributes. The label on each piece of data is dynamically weighted, which is based on the ``confidence value'' of each robot in each piece of data. We define robots as actors in reinforcement learning. Different robots or the same robot in different environments are different actors.

The ``confidence value'' motioned above of the actor is the degree of confirmation on which action the robot chooses to perform. For example, in a piece of sample from training data, the private Network 1 evaluates Q-values of different actions to (85, 85, 84, 83, 86), but the evaluation of the k-G sharing network is (20, 20, 100, 10, 10). In this case, we are more confident on actor of k-G sharing network, because it has significant differentiation in the scoring process. On the contrary, the scores from actor of private Network 1 are confusing. Therefore, when generating the labels, the algorithm calculates the confidence value according to the score of different actors. Then the scores are weighted with confidence value and summed up. Finally, we obtain labels of training data by executing the above steps for each piece of data. There are several approaches to define confidence, such as variance, standard deviation, and information entropy. From the definition of the above statistical indicators we can infer that using variance to describe uncertainty will fail in some cases because it requires uniform distribution of data and ignores the occurrence of extreme events. The variance needs to meet the relevant premise to describe the uncertainty of the information. Entropy is more suitable for describing the uncertainty of information than variance, which comes from the definition of entropy. Uncertainty is the embodiment of confidence. So, In this work, we use information entropy to define confidence. Formula (1) is quantitative function of robotic confidence (information entropy):

Robot j "confidence":
$$
c _ { j } = - \frac { 1 } { \ln m } \sum _ { i = 1 } ^ { m } \left( \frac { s c o r e _ { i j } } { \sum _ { i = 1 } ^ { m } s c o r e _ { i j } } \cdot \ln \left( \frac { s c o r e _ { i j } } { \sum _ { i = 1 } ^ { m } s c o r e _ { i j } } \right) \right)\eqno{(1)}
$$
m is the action size of robot, n is the number of private networks. 
Memory weight of robot j:
$$
w _ { j } = \frac { \left( 1 - c _ { j } \right) } { \sum _ { j = 1 } ^ { n } \left( 1 - c _ { j } \right) } \eqno{(2)}
$$
Knowledge fusion function:
$$
label_ { j } = \text { score } \times \left( c _ { 1 } , c _ { 2 } , \cdots c _ { m } \right) ^ { T }\eqno{(3)}
$$
It should be noted that Fig.4 only shows the process of one sample generating one label. Actually, we need to generate a large number of samples. For each data sample, the confidence values of the actors are different, so the weight of each actor is not the same. For example, when we generate 50,000 different pieces of data, there are nearly 50,000 kinds of different combinations of confidence. These changing weights can be incorporated into the data labels, and enable the generated network to dynamically adjust the weights on different sensor data. In conclusion, knowledge fusion algorithm in cloud can be defined as:
$$
\omega _ { j } = \left( 1 - c _ { j } \right) \div \sum _ { j = 1 } ^ { n } \left( 1 - c _ { j } \right)\eqno{(4)}
$$

$$
y _ { i } = \sum _ { j = 1 } ^ { num } c _ { j } \cdot \omega _ { j }\eqno{(5)}
$$

$$
L \left( y , h _ { \theta } \left( x _ { i } \right) \right) = \frac { 1 } { N } \sum _ { i = 1 } ^ { N } \left( y _ { i } - h _ { \theta } \left( x _ { i } \right) \right) ^ { 2 }\eqno{(6)}
$$

$$
\theta ^ { * } = \arg \min _ { \theta } \frac { 1 } { N } \sum _ { i = 1 } ^ { N } L \left( y _ { i } \cdot h _ { \theta } \left( x _ { i } \right) \right)\eqno{(7)}
$$

Formula 4 takes the proportion of the confidence of robots as a weight. Formula 5 obtains the label of the sensor data by weighted summation. Formula 6 defines the error in the training process of generative network. Formula 7 is the goal of the training process.
We discribe our approach in details in Algorithm 2. For a single robot, private network is obtained in different environments. Therefore, it can be regarded as asynchronous learning of the robot in LFRL. When there are multiple robots, we just need to treat them as the same robot in different environments. At this time, the evolution process is asynchronous, and multiple robots are synchronized.

\begin{algorithm}
	\caption{Knowledge Fusion Algorithm}
	Initialize the shared network with random Parameters ${\theta }$ \;
	\KwIn{$K$: The number of data samples generated ; $N$: The number of private networks;  $M$: Action sizes of the robot; } 
	\KwOut{$\theta$} 
	\For{i=1,i $\le$ N,i++}{
		$\hat{x_i}$ $\leftarrow$  Calculate indirct features from $\tilde{x_i}$\;
		$x_i$ $\leftarrow$ [$\tilde{x_i},\hat{x_i}$] \;
		\For{n=1,n $\le$ K,n++}{
			$score_{in}$ $\leftarrow$ $f_n(x_i)$\;
			$score_{i}$ $\leftarrow$ score append $score_{in}$
		}
		\For{n=1,n $\le$ K,n++}{
			\For{m=1,m $\le$ M,m++}{
				$c_{in}$ $\leftarrow$ Calculate the confidence value of the n-th private network in the i-th data based on formula (1)
			}
		}

		$label_i$ $\leftarrow$ Calculate the $label_i$ based on formula (2) and (3)\;
		$label$ $\leftarrow$ label append $label_i$\;
	}
	$\theta$ $\leftarrow$ training the shared network from (x,label)\;
\end{algorithm}
It should be explained that the shared model in the cloud is not the final policy model of the local robot. We only use the shared model in the cloud as a pre-trained model or a feature extractor. The shared model maintained in the cloud is a cautious policy model but not the optimal for every robot. That is to say, the shared model in the cloud will not make serious mistakes in some private unstructured environments but the action is not the best. It is necessary for the robot to train its own policy model based on the shared model from the cloud, otherwise the error rate will be high in its private unstructured environment. As the saying goes, the older the person, the smaller the courage. In the process of lifelong learning, the cloud model will become more and more “timid”. In order to remove the error rate, we should transfer the shared model and train a new private model through reinforcement learning in a new environment. This responded to the transfer learning process in LFRL.
\subsection{Transfer the shared model}
Various approaches of transfer reinforcement learning have been proposed. In the specific task that a robot learns to navigate, we found that there are two applied approaches. The first one is taking the shared model as initial actor network. While the other one is using the shared model as a feature extractor.
If we adopt the first approach that takes the shared model as an initial actor network, abilities of avoiding obstacles and reaching targets can remain the same. In this approach, the robot deserves a good score at the beginning. The experimental data shows that the final score of the robot has been greatly improved at the same time. However, every coin has two sides, this approach is unstable. The training time depends on the adjustment of parameters in some extent. For example, we should accelerate updating speed, increase the punishment, reduce the probability of random action etc.

The shared model also can be used as a feature extractor in transfer reinforcement learning. As illustrated in Fig.5, this method increases the dimension of the features. So, it can improve the effect stably. One problem that needs to be solved in experiment is that there is a structural difference between input layer of the shared network and private network. The approach in LFRL is that the number of nodes in the input layer is consistent with the number of elements in the original feature vector, as shown in Fig.5. The features from transfer learning are not used as inputs to the shared network. They are just inputs of training private networks. This approach has high applicability, even though the shared model and private models have different network structures.

It is also worth noting that if the robot uses image sensors to acquire images as feature data. It is recommended to use the traditional transfer learning method that taking the output of some convolutional layers as features because the Q-network is a convolutional neural network. If a non-image sensor such as a laser radar is used, the Q-network is not a convolutional neural network, then we will use the output of the entire network as additional features, as Fig.5 shows.
\begin{figure}[thpb]
	\centering
	\includegraphics[width=1\linewidth]{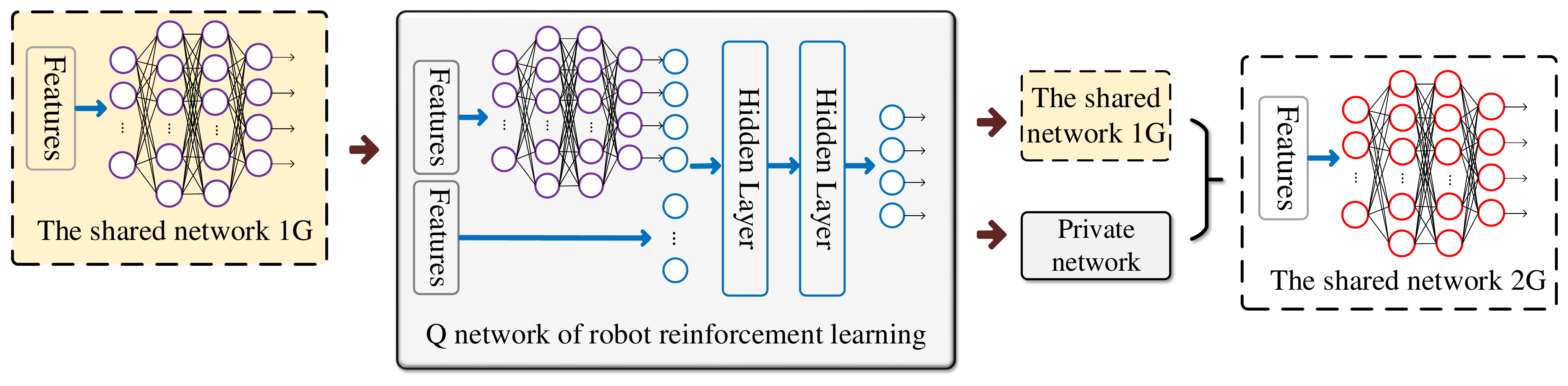}
	\caption{A transfer learning method of LFRL}
	\label{fig:architecture}
\end{figure}
The learning process of the robot can roughly divided into two stages, the stage of avoiding obstacles and the stage of reaching the target. With the former transfer algorithm, it is possible to utilize the obstacle avoidance skills of the cloud model. In the latter transfer method, it is obvious that the evaluation of the directions by cloud model is valuable, which is useful features for navigation.
\subsection{Explanation from human cognitive science}
The design of the LFRL is inspired by the human decision-making process in cognitive science. For example, when playing chess, the chess player will make decisions based on the rules and his own experiences. The chess experiences include his own experiences and the experiences of other chess players he has seen. We can regard the chess player as a decision model. The quality of the decision model represents the performance level of the chess player. In general, this policy model will become increasingly excellent through experience accumulation, and the chess player's skill will be improved. This is the iteratively evolutionary process in LFRL. After each chess player finishing playing chess, his chess level or policy model evolves, which is analogous to the process of knowledge fusion in LFRL. And these experiences will also be used in later chess player, which is analogous to the process of transfer learning in LFRL. 
Fig. 1 demonstrates a concrete example. The person on the right is considering where should the next step goes. The chess he has played and the chess he has seen are the most two influential factors on making decision. But his chess experiences may influence the next step differently. At this time, according to human cognitive science, the man will be more influenced by experiences with clear judgments. An experience with a clear judgment will have a higher weight in decision making. This procedure of humans makes decisions is analogous to knowledge fusion algorithm in LFRL. The influence of different chess experience is always dynamic in the decision of each step. The knowledge fusion algorithm in LFRL achieves this cognitive phenomenon by adaptively weighting the labels of training data. The chess player is a decision model that incorporates his own experiences. Corresponding to this opinion, LFRL integrates experience into one decision model by generating a network. This process is also analogous to the operation of human cognitive science.
\section{Experiments}
In this section, we intend to answer three questions: 1) Can LFRL help reduce training time without sacrifice accuracy of navigation in cloud robotic systems? 2) Does the knowledge fusion algorithm is effective to increase the shared model? 3) Are transfer learning approaches effective to transfer the shared model to specific task? To answer the former question, we conduct experiments to compare the performance of the generic approach and LFRL. To answer the second question, we conduct experiments to compare the performance of generic models and the shared model in transfer reinforcement learning. To answer the third question, we conduct experiments to compare the performance of the two transfer learning approaches and the native reinforcement learning.
\begin{figure*}[thpb]
	\centering
	\subfigure[Env-1]{
		\label{fig:subfig:a} 
		\includegraphics[width=0.22\linewidth]{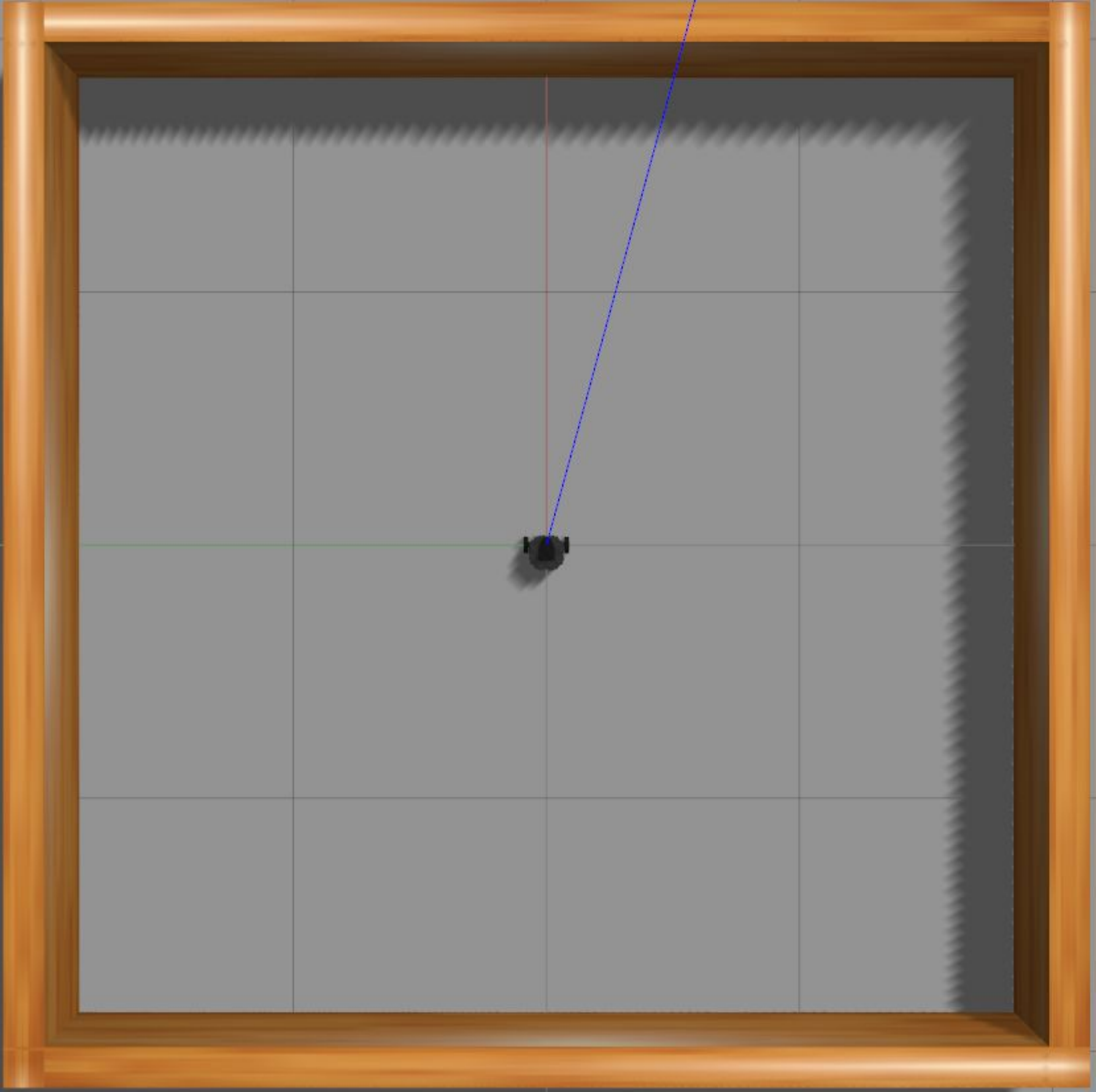}}
	\subfigure[Env-2]{
		\label{fig:subfig:b} 
		\includegraphics[width=0.22\linewidth]{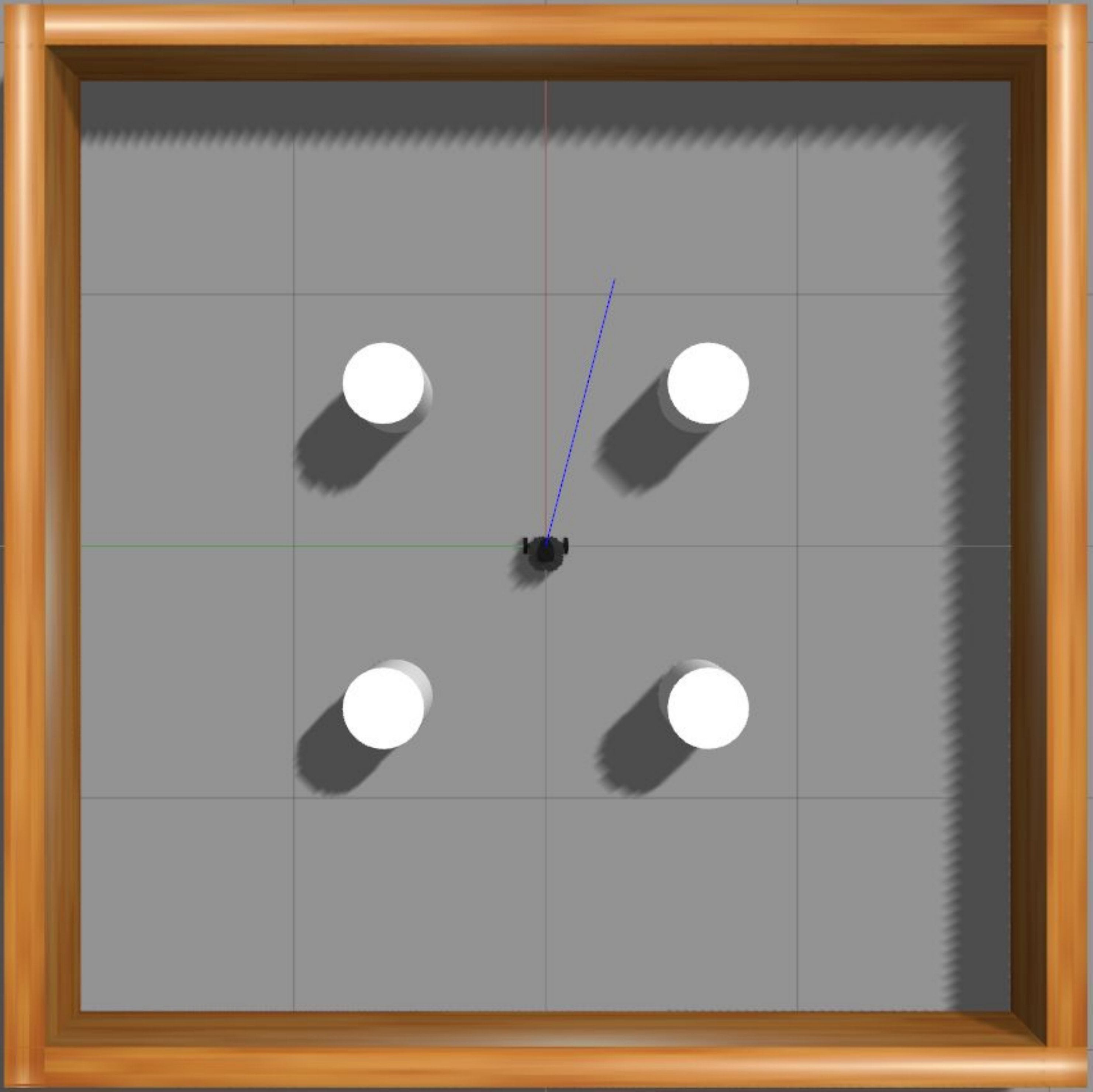}}
	\subfigure[Env-3]{
		\label{fig:subfig:b} 
		\includegraphics[width=0.22\linewidth]{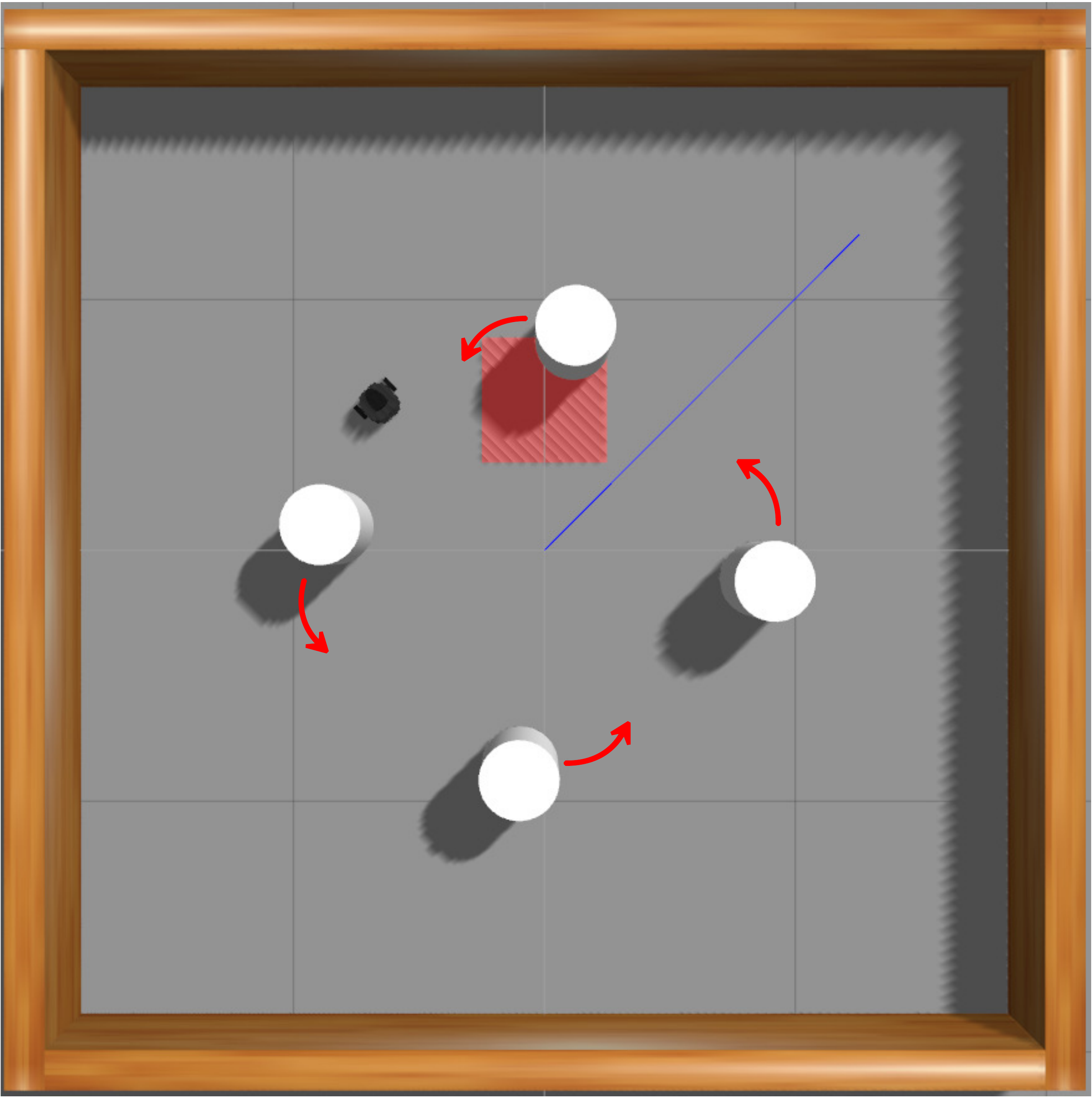}}
	\subfigure[Env-4]{
		\label{fig:subfig:b} 
		\includegraphics[width=0.22\linewidth]{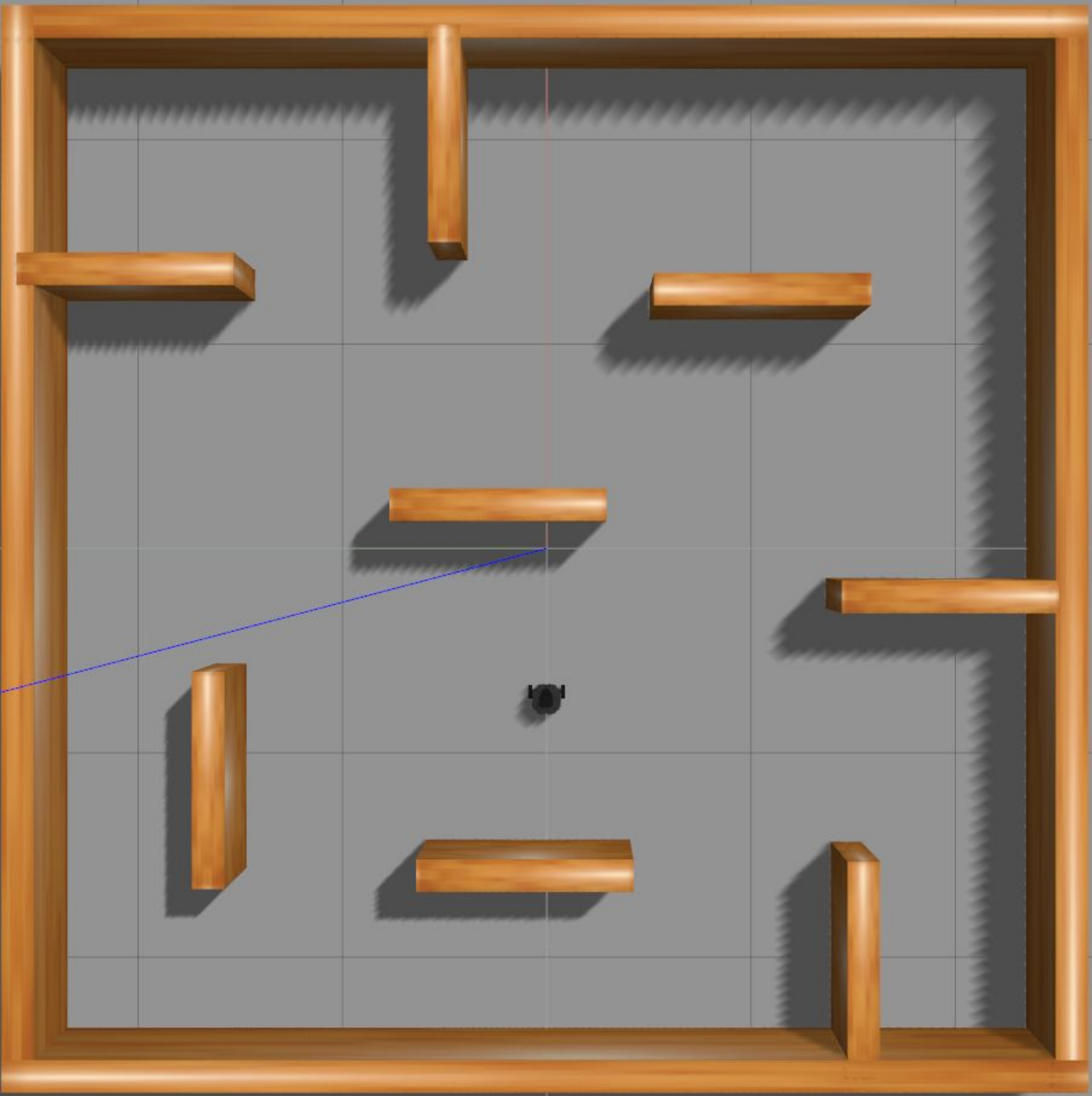}}
	\subfigure[Env-1 score]{
		\label{fig:subfig:b} 
		\includegraphics[width=0.22\linewidth]{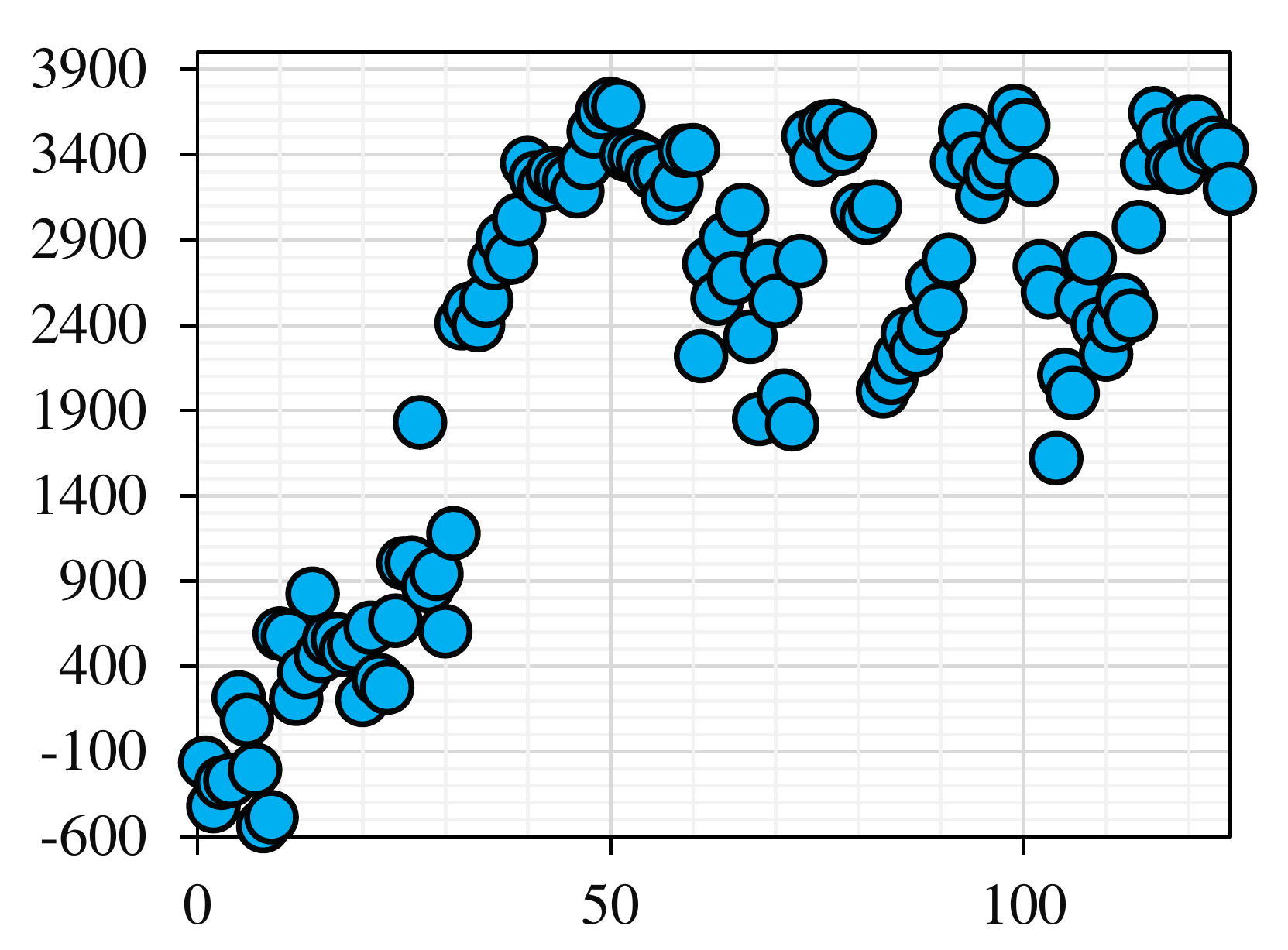}}
	\subfigure[Env-2 score]{
		\label{fig:subfig:b} 
		\includegraphics[width=0.22\linewidth]{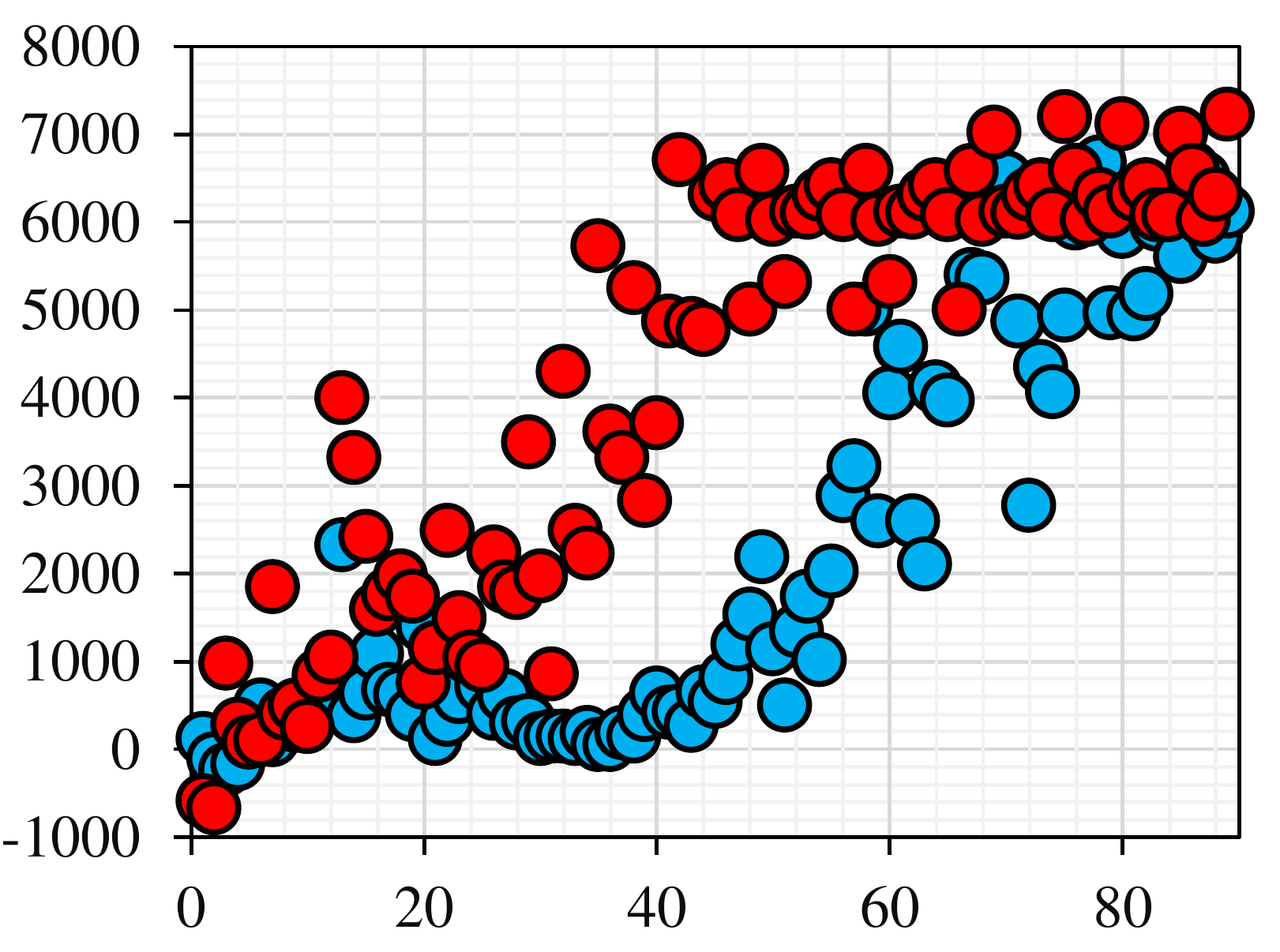}}
	\subfigure[Env-3 score]{
		\label{fig:subfig:b} 
		\includegraphics[width=0.22\linewidth]{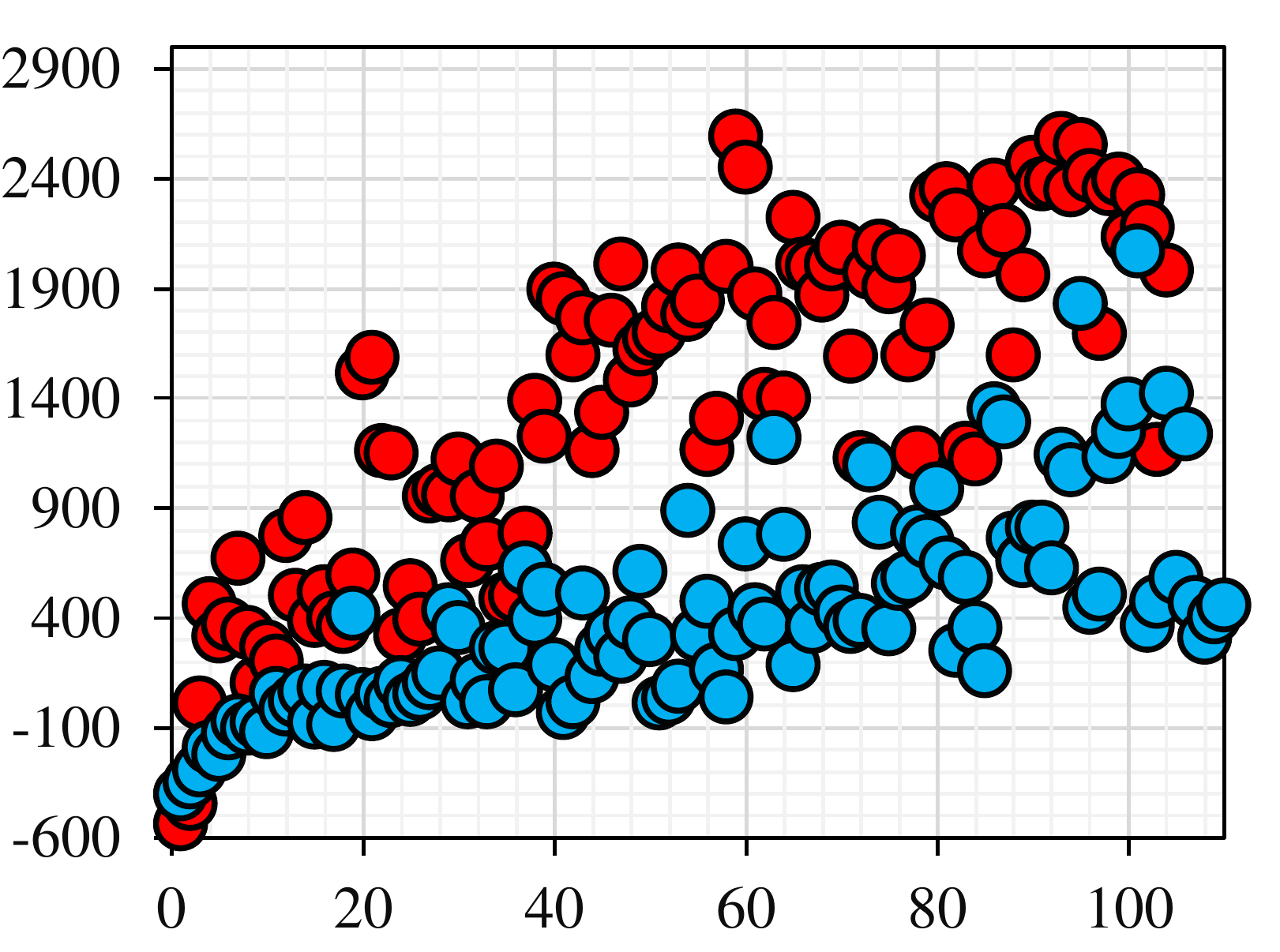}}
	\subfigure[Env-4 score]{
		\label{fig:subfig:b} 
		\includegraphics[width=0.22\linewidth]{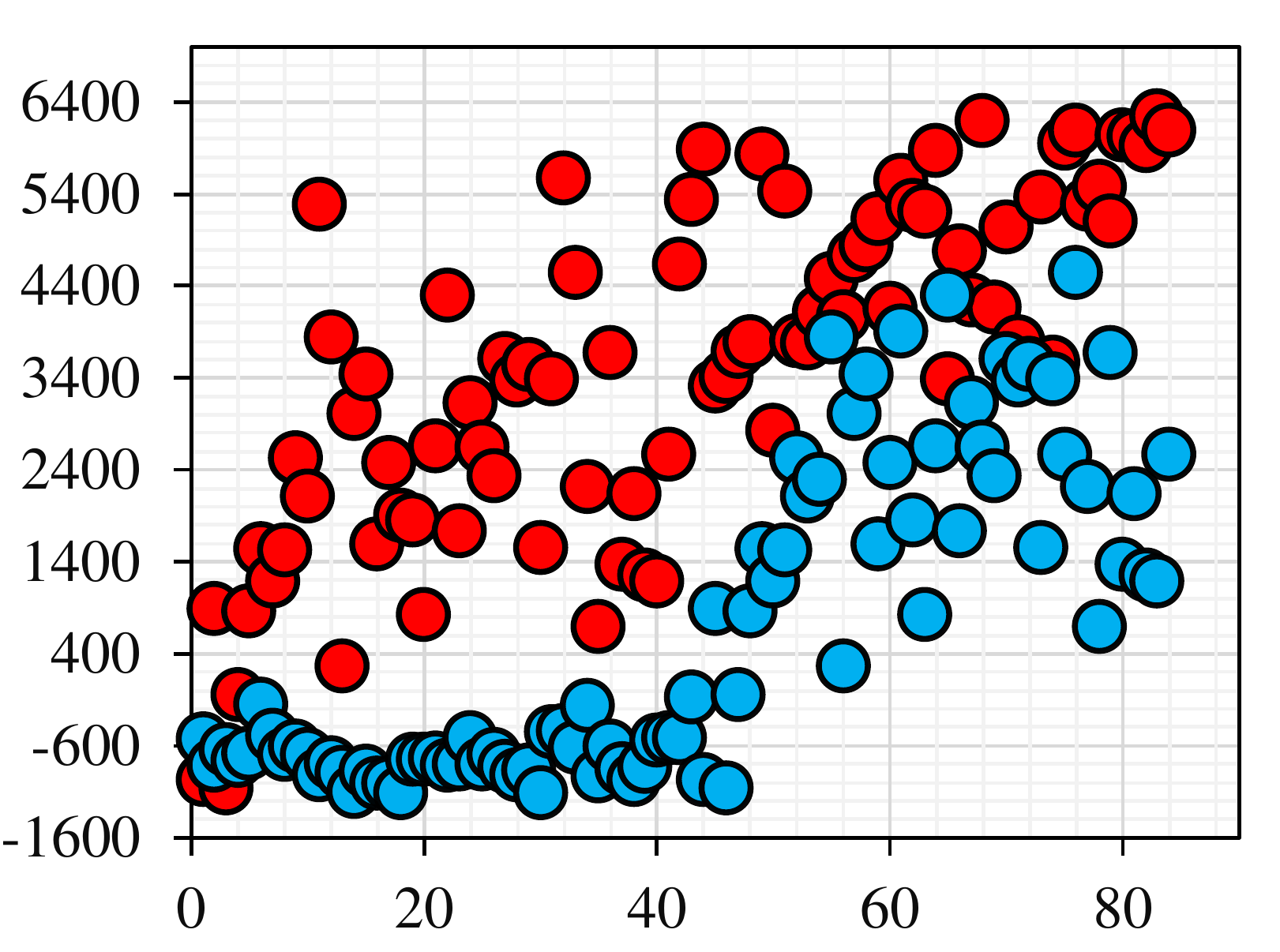}}
	\caption{We present both quantitative and compared results: Subfigure a to Subfigure d are the training environments. Subfigure e to h present scores of the generic approach (blue) compared with LFRL approach (red) in training process.In the training procedure of Env-1, LFRL has the same result with generic approaches. Because there is no antecedent shared models for the robot. In the training procedure of Env-2, LFRL obtained the shared model 1G, which made LFRL get higher reward in less time compared with the generic approach. In Env-3 and Env-4, LFRL evolve the shared model to 2G and 3G and obtained excellent result. From this figure, we demonstrate that LFRL can get higher reward in less time compared with the generic approach.}
	\label{fig:subfig} 
\end{figure*}
\subsection{Experimental setup}
The training procedure of the LFRL was implemented in virtual environment simulated by gazebo. Four training environments were constructed to show the different consequence between the generic approach training from scratch and LFRL, as shown in Fig.6. There is no obstacle in Env-1 except the walls. There are four static cylindrical obstacles in Env-2, four moving cylindrical obstacles in Env-3. More complex static obstacles are in Env-4. In every environment, a \textit{Turtlebot3} equipped with a laser range sensor is used as the robot platform. The scanning range is from 0.13m to 4m. The target is represented by a red square object. During the training process, the starting pose of the robot is the geometric center of the ground in the training environment. The target is randomly generated in pose where there are no obstacles in the environment. An episode is terminated after the agent either reaches the goal, collides with an obstacle, or after a maximum of 6000 steps during training and 1000 for testing. We calculated the average reward of the robot every two minutes. We end the training when the average reward of the agent is enough and stable. We trained the model from scratch on a single Nvidia GeForce GTX 1070 GPU. The actor-critic network used two fully connected layers with 64 units. The output is used to produce the discrete action probabilities by a linear layer followed by a softmax, and the value function by a linear layer.
\begin{figure*}[thpb]
	\centering
	\subfigure[Env-1]{
		\label{fig:subfig:a} 
		\includegraphics[width=0.2\linewidth]{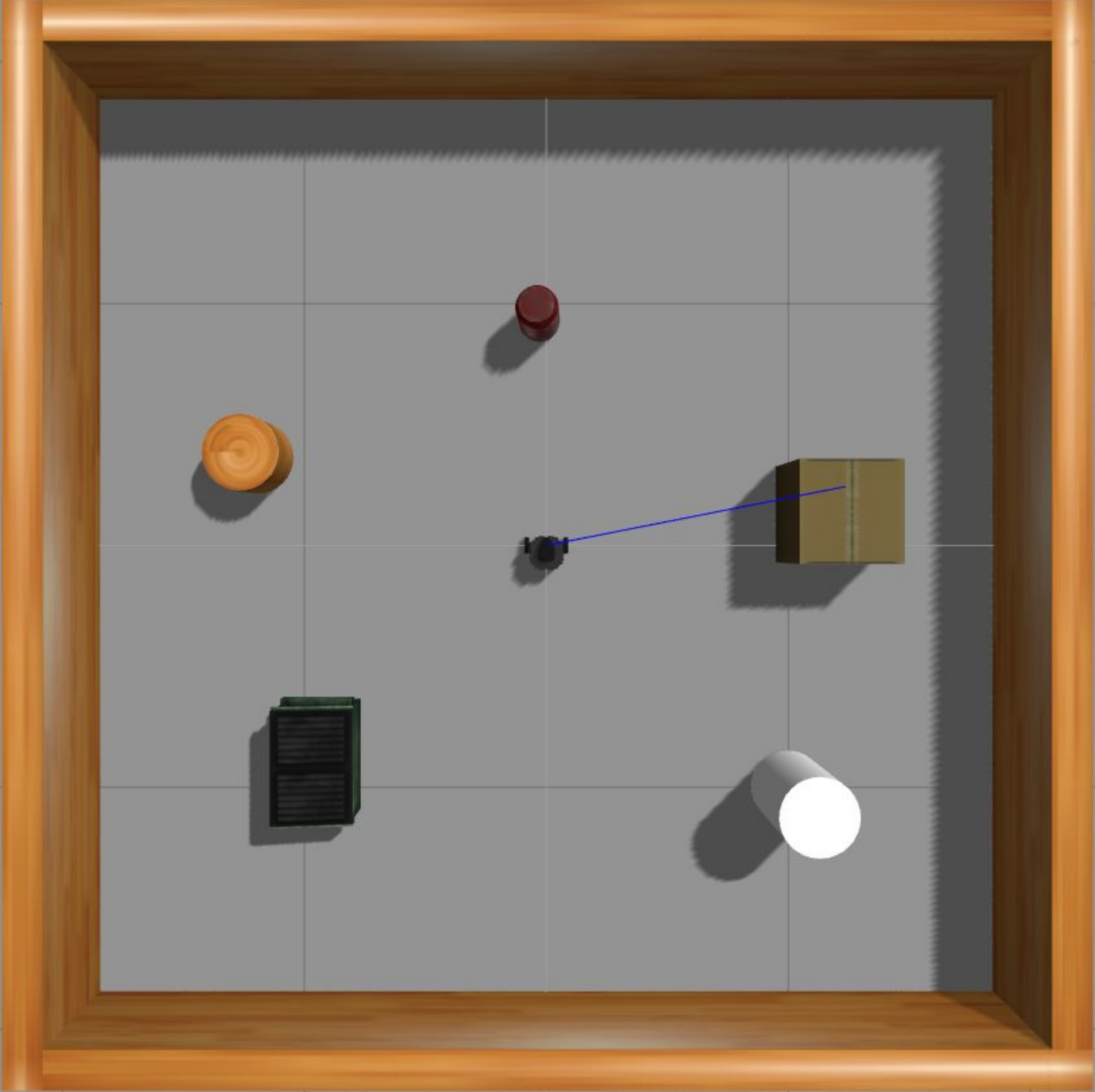}}
	\hspace{0.2in}
	\subfigure[Env-2]{
		\label{fig:subfig:b} 
		\includegraphics[width=0.2\linewidth]{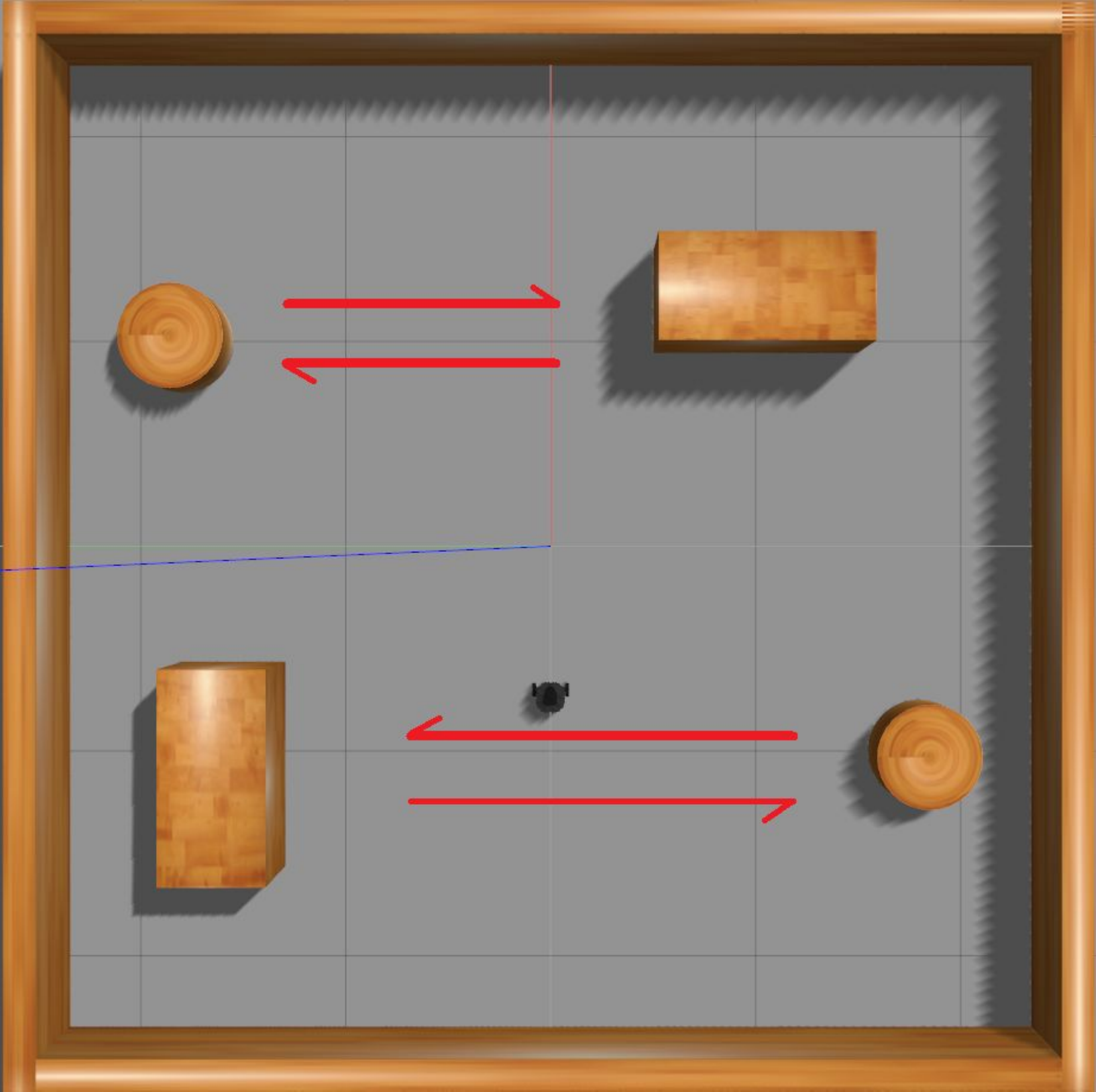}}
	\hspace{0.2in}
	\subfigure[Env-3]{
		\label{fig:subfig:b} 
		\includegraphics[width=0.2\linewidth]{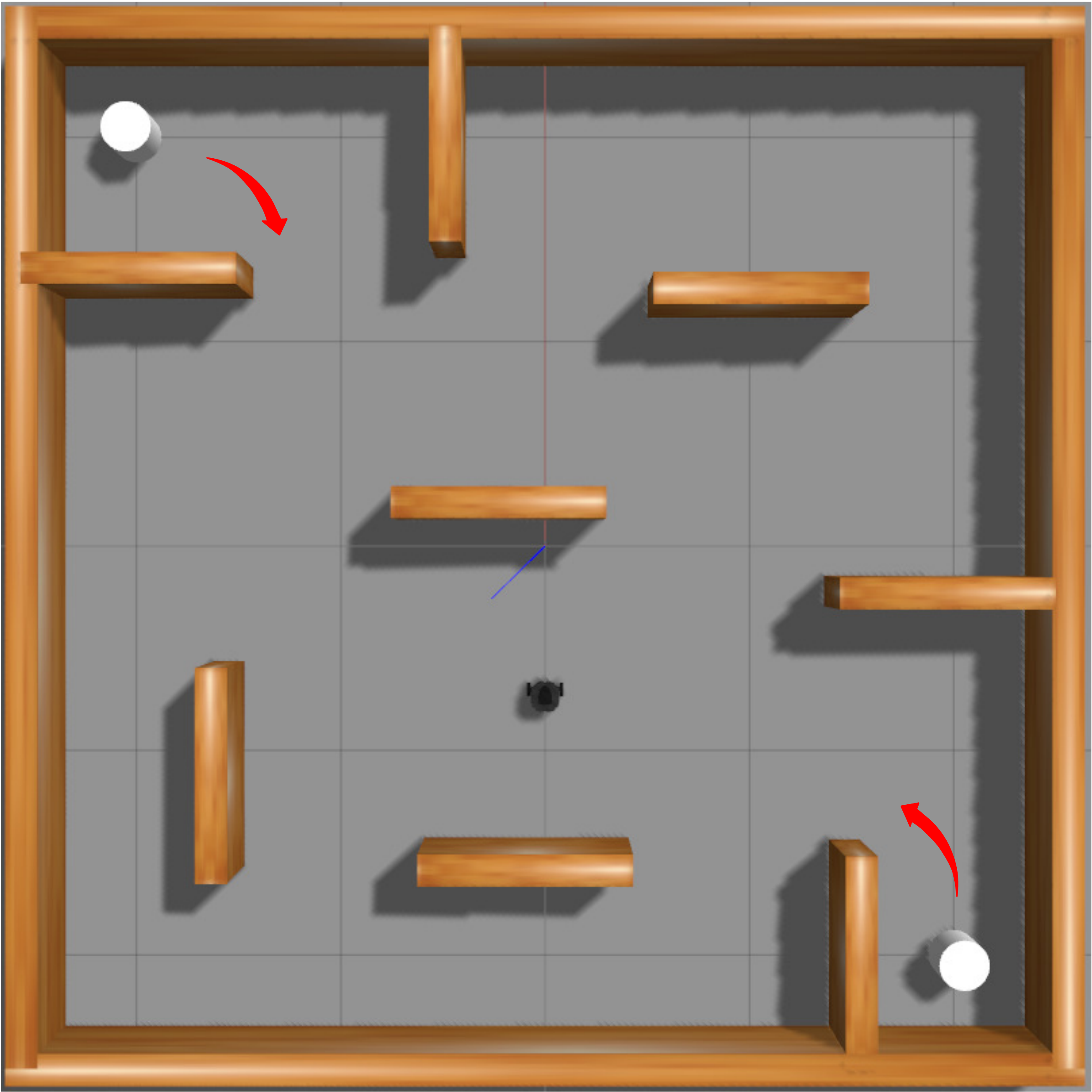}}
	\subfigure[Env-1 stacked scores]{
		\label{fig:subfig:b} 
		\includegraphics[width=0.31\linewidth]{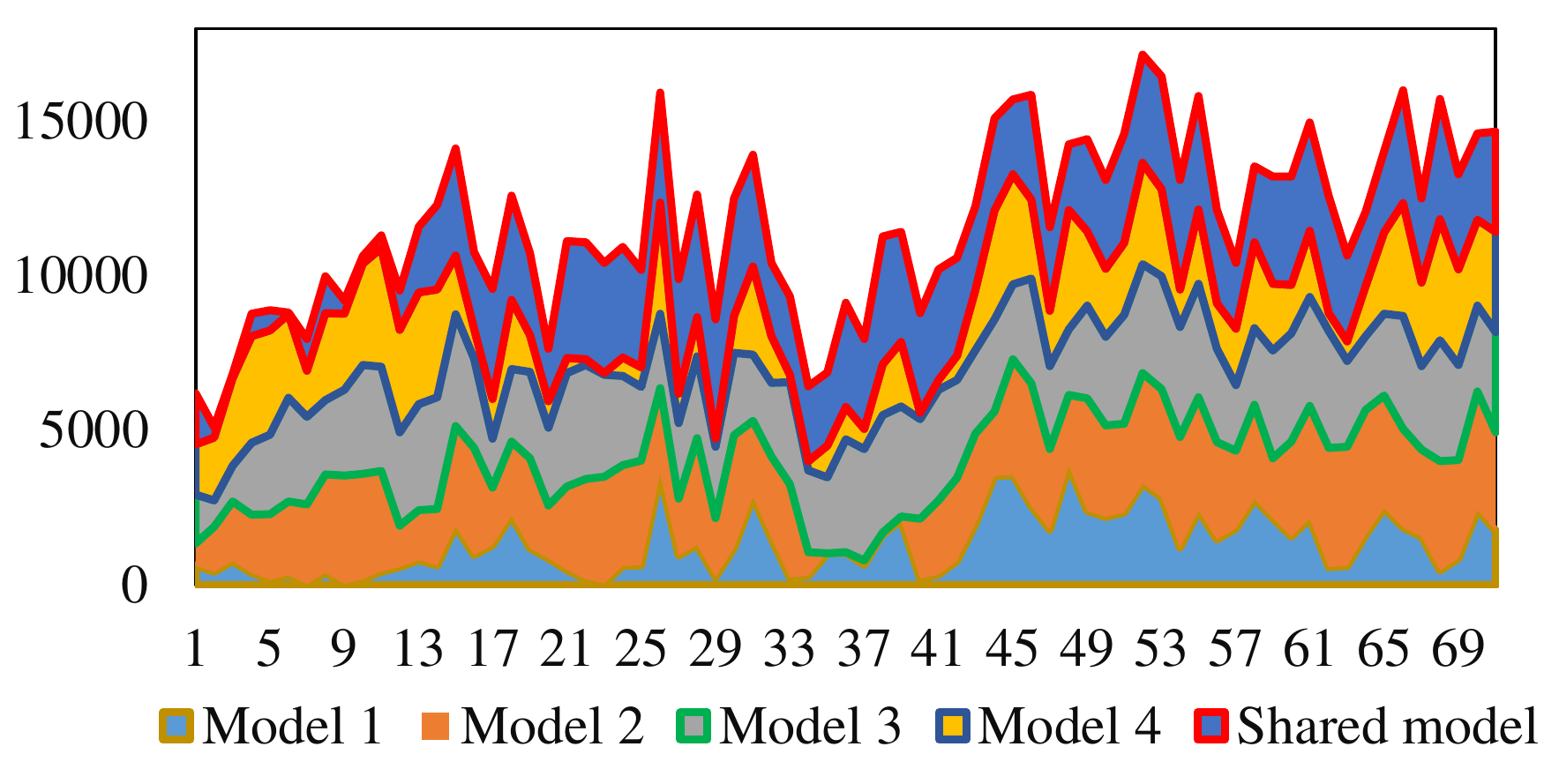}}
	\subfigure[Env-2 stacked scores]{
		\label{fig:subfig:b} 
		\includegraphics[width=0.3\linewidth]{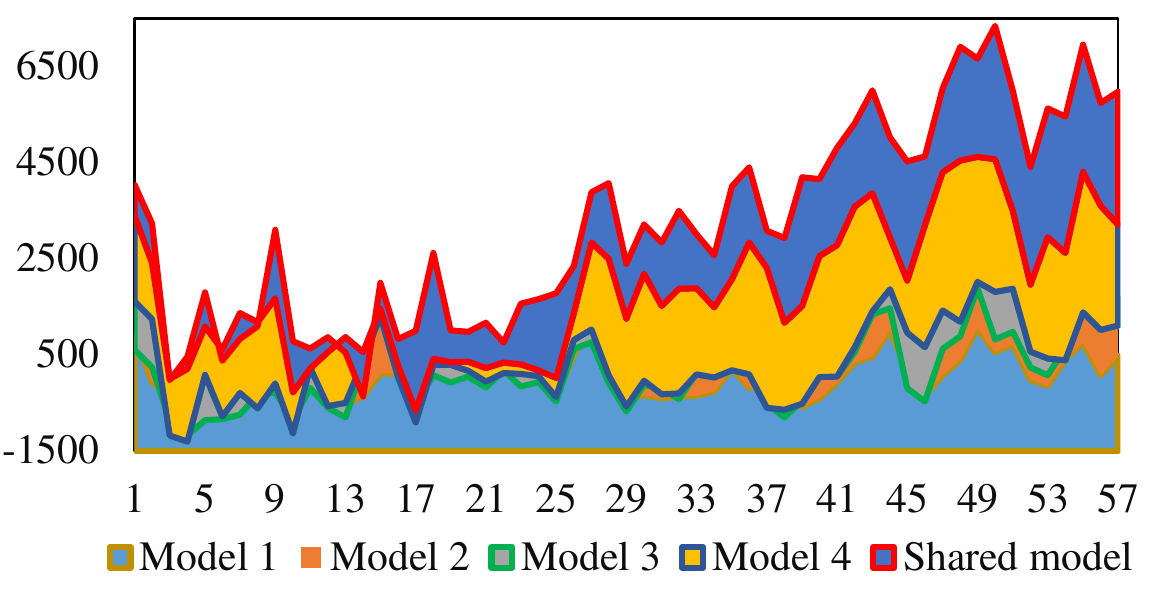}}
	\subfigure[Env-3 stacked scores (sampling)]{
		\label{fig:subfig:b} 
		\includegraphics[width=0.3\linewidth]{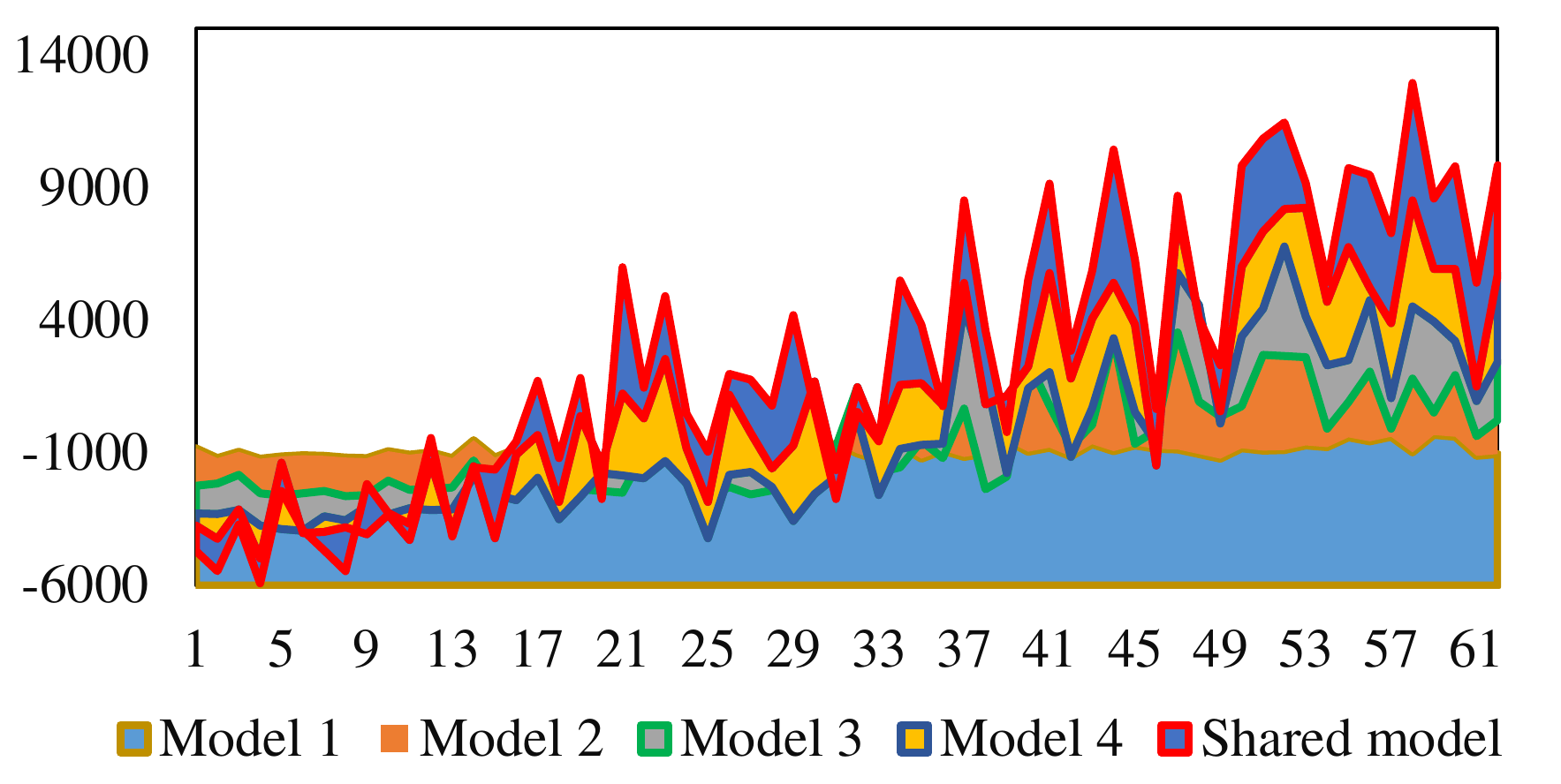}}
	\caption{We present both quantitative and compared results: Subfigure a to Subfigure c are the testing environments. Because of the large amount of data, we present the stacked figure to better repersent the comparison. From the subfigure d to subfigure f, it can seen that the reward of the shared model accounts for a larger proportion in all positive rewards (greater than 0) models in the later stage of training.}
	\label{fig:subfig} 
\end{figure*}
\begin{table*}[htbp]
	\centering
	\caption{Results of the contrast experiment}
	\begin{tabular}{cccccccccc}
		\toprule
		\multirow{2}[4]{*}{} & \multicolumn{3}{c}{Time to meet requirement} & \multicolumn{3}{c}{Averange scores} & \multicolumn{3}{c}{Averange of the last five scores} \\
		\cmidrule{2-10}          & Test-Env-1 & Test-Env-2 & Test-Env-3 & Test-Env-1 & Test-Env-2 & Test-Env-3 & Test-Env-1 & Test-Env-2 & Test-Env-3 \\
		\midrule
		model 1 & \cellcolor[rgb]{ .949,  .949,  .949}1h 32min & $ > $3h   & $ > $6 h  & 1353.36 & 46.5  & -953.56 & 1421.54 & \cellcolor[rgb]{ .949,  .949,  .949}314.948 & -914.16 \\
		\midrule
		model 2 & \cellcolor[rgb]{ .651,  .651,  .651}33min & $ > $3h   & $ > $6 h  & \cellcolor[rgb]{ .851,  .851,  .851}2631.57 & \cellcolor[rgb]{ .949,  .949,  .949}71.91 & \cellcolor[rgb]{ .949,  .949,  .949}288.61 & \cellcolor[rgb]{ .651,  .651,  .651}3516.34 & \cellcolor[rgb]{ .851,  .851,  .851}794.02 & \cellcolor[rgb]{ .949,  .949,  .949}-132.07 \\
		\midrule
		model 3 & \cellcolor[rgb]{ .749,  .749,  .749}37min & $ > $3h   & \cellcolor[rgb]{ .851,  .851,  .851}5h 50min & \cellcolor[rgb]{ .651,  .651,  .651}2925.53 & \cellcolor[rgb]{ .851,  .851,  .851}166.5 & \cellcolor[rgb]{ .851,  .851,  .851}318.04 & \cellcolor[rgb]{ .851,  .851,  .851}3097.64 & -244.17 & \cellcolor[rgb]{ .851,  .851,  .851}1919.12 \\
		\midrule
		model 4 & 4h 41min & \cellcolor[rgb]{ .749,  .749,  .749}2h 30min & \cellcolor[rgb]{ .749,  .749,  .749}5h 38min & \cellcolor[rgb]{ .949,  .949,  .949}1989.51 & \cellcolor[rgb]{ .651,  .651,  .651}1557.24 & \cellcolor[rgb]{ .749,  .749,  .749}1477.5 & \cellcolor[rgb]{ .949,  .949,  .949}2483.18 & \cellcolor[rgb]{ .749,  .749,  .749}2471.07 & \cellcolor[rgb]{ .749,  .749,  .749}3087.83 \\
		\midrule
		Shared model & \cellcolor[rgb]{ .851,  .851,  .851}55min & \cellcolor[rgb]{ .651,  .651,  .651}2h 18min & \cellcolor[rgb]{ .651,  .651,  .651}4h 48min & \cellcolor[rgb]{ .749,  .749,  .749}2725.16 & \cellcolor[rgb]{ .749,  .749,  .749}1327.76 & \cellcolor[rgb]{ .651,  .651,  .651}1625.61 & \cellcolor[rgb]{ .749,  .749,  .749}3497.66 & \cellcolor[rgb]{ .651,  .651,  .651}2617.02 & \cellcolor[rgb]{ .651,  .651,  .651}3670.92 \\
		\bottomrule
	\end{tabular}%
	\begin{tablenotes}
		\footnotesize
		\item[1] The background color of the cell reflects the performance of the corresponding model. The darker the color, the better the performance.
	\end{tablenotes}
	\label{tab:addlabel}%
\end{table*}%
\subsection{Evaluation for the architecture}
To show the performance of LFRL, we tested it and compared with generic methods in the four environments. Then we started the training procedure of LFRL. As mentioned before, we initialized the shared model and evolved it as Algorithm 2 after training in Env-1. In the cloud robotic system, the robot downloaded the shared model 1G. Then, the robot performed reinforcement learning based on the shared model. The robot got a private model after training and it would be uploaded to the cloud server. The cloud server fused the private model and the shared model 1G to obstain the shared model 2G. With the same mode, follow-up evolutions would be performed. We constructed four environments, so the shared model upgraded to 4G. Performance of LFRL shown in Fig.6 where also shows generic methods performance. In Env2-Env4, LFRL increased accuracy of navigating decision and reduced training time in the cloud robotic system. From the last row of Fig.6, we can observe that the improvement are more efficient with the shared model. LFRL is highly effective for learning a policy over all considered obstacles. It improves the generalization capability of our trained model across commonly encountered environments. Experiments demonstrate that LFRL is capable of reducing training time without sacrificing accuracy of navigating decision in cloud robotic systems.
\subsection{Evaluation for the knowledge fusion algorithm}
In order to verify the effectiveness of the knowledge fusion algorithm, we conducted a comparative experiment. We created three new environments that were not present in the previous experiments. These environments are more similar to real situations: Static obstacles such as cardboard boxes, dustbin, cans are in the Test-Env-1. Moving obstacles such as stakes are in Test-Env-2. Test-Env-3 includes more complex static obstacles and moving cylindrical obstacles. We still used the \textit{Turtlebot3} created by gazebo as the testing platform. In order to vertify the advancement of the shared model, we trained the navigation policy based on the generic model 1, model 2, model 3, model 4 in Test-Env-1, Test-Env-2 and Test-Env-3 respectively. The generic models are from the previous generic approaches experiments. These policy models are trained from one environment without knowledge fusion. According to the hyper-parameters and complexity of environments, the average score goal (5 consecutive times above a certain score) in Test-Env-1 is 4000, Test-Env-2 is 3000, Test-Env-3 is 2600.

In the following, we present compared results in Fig.7 and quantitative results in Table 1. The shared model steadily reduces training time. In particular, we can observe that the generic method models are only able to make excellent decisions in individual environments; while the shared model is able to make excellent decisions in plenty of different environments. So, the proposed knowledge fusion algorithm in this paper is effective.
\subsection{Evaluation for the two transfer learning approaches}
In order to verify and compare the two transfer learning approaches, we conducted a comparative experiment. The result is present in Fig.8. It can be seen from the figure that both transfer learning approaches can effectively improve the efficiency of reinforcement learning. Among them, the approach of parameter transferring has faster learning speed and the approach of feature extractor has higher stability.
\begin{figure}[thpb]
	\centering
	\includegraphics[width=0.8\linewidth]{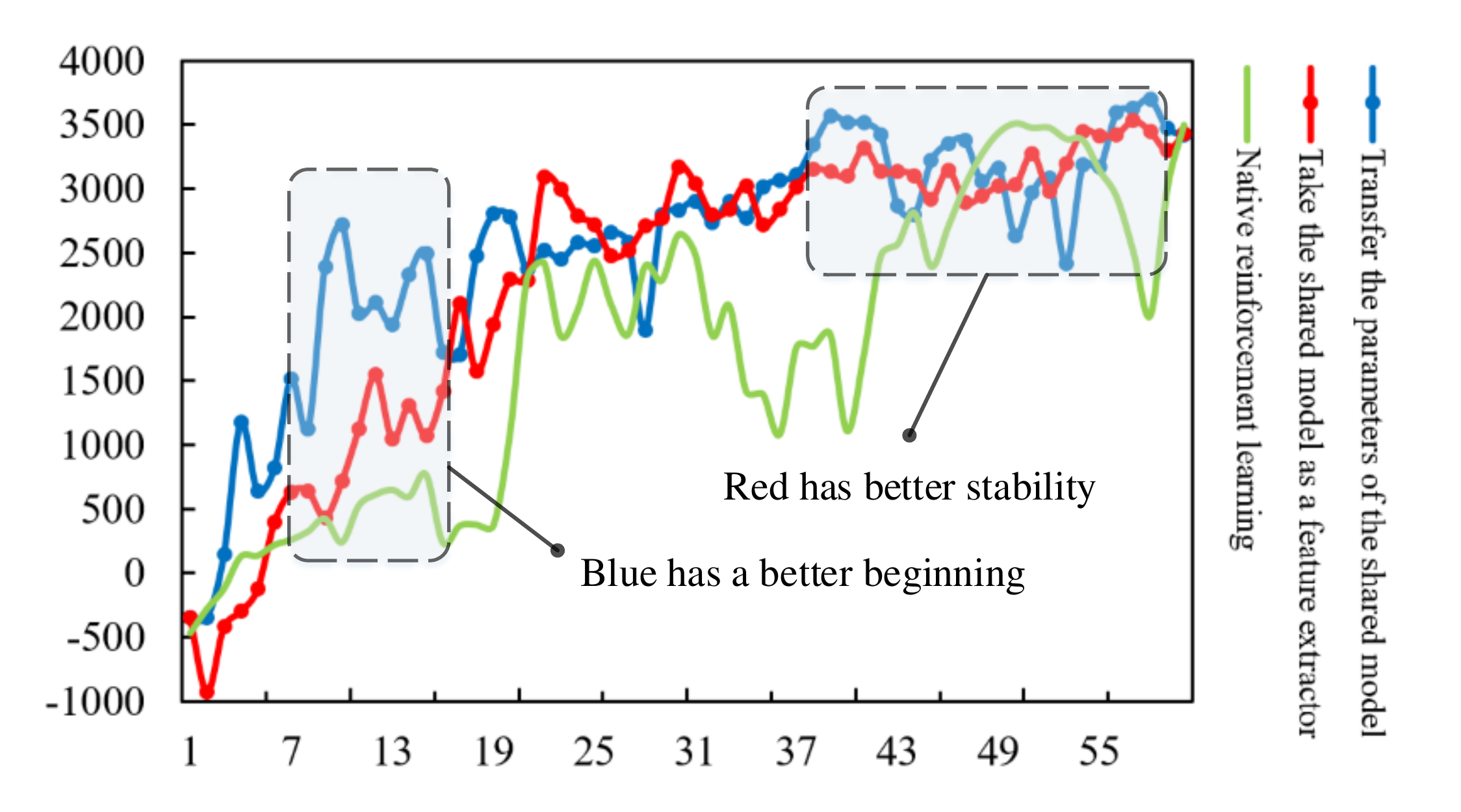}
	\caption{Transfer learning approaches comparison}
	\label{fig:architecture}
\end{figure}
\subsection{Real world experiments}
\begin{figure}[thpb]
	\centering
	\includegraphics[width=0.6\linewidth]{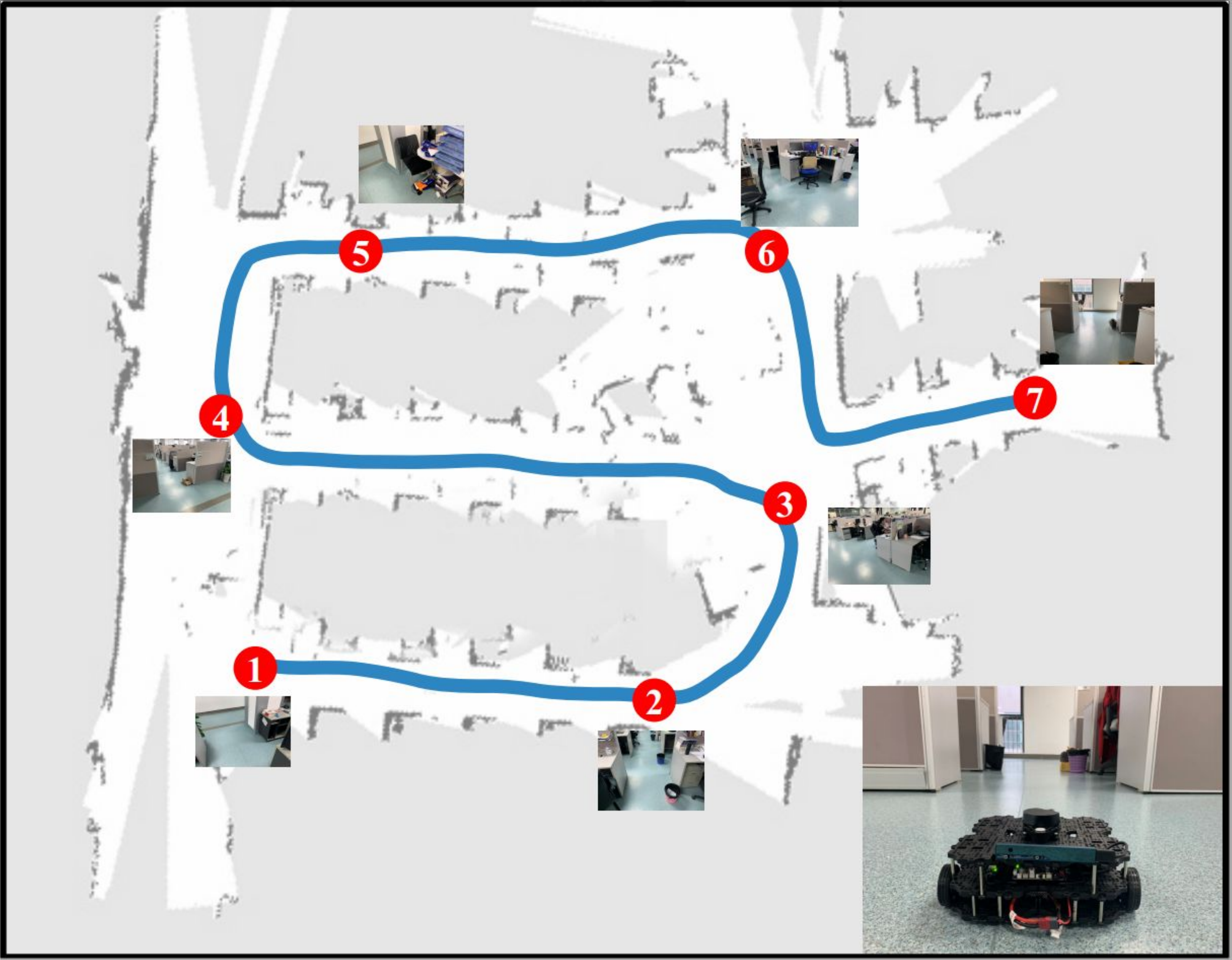}
	\caption{Trajectory tracking in the real environment}
	\label{fig:architecture}
\end{figure}
We also conducted real-world experiments to test the performance of our approach with different sensor noise. We use Turtlebot 3 platform, which is shown in Fig.9. The shared model 4G in the cloud was downloaded. Then we performed transferring reinforcement learning in Env-4 and got the policy finally. The Turtlebot navigates automatically in an indoor office environment as shown in Fig.9 under the policy. The experiment indicates that the policy is reliable in real environment. The reference [10] also corroborates the conclusion.
\section{conclusion}
We presented a learning architecture LFRL for navigation in cloud robotic systems. The architecture is able to make navigation-learning robots effectively use prior knowledge and quickly adapt to new environment. Additionally, we presented a knowleged fusion algorithm in LFRL and introduced transfer methods. Our approach is able to fuse models and asynchronously evolve the shared model. We validated our architecture and algorithmes in policy-learning experiments and realsed a website to provide the service.

The architecture has fixed requirements for the dimensions of input sensor signal and the dimensions of action. We leave it as future work to make LFRL flexible to deal with different input and output dimensions. The more flexible LFRL will offer a wider range of services in cloud robotic systems.
\section{Acknowledgement}
This work was supported by National Natural Science Foundation of China No. 61603376; Guangdong-Hongkong Joint Scheme (Y86400); Shenzhen Science, Technology and Innovation Commission (SZSTI) Y79804101S awarded to Dr. Lujia Wang. The work also inspired and supported by Chengzhong Xu from University of Macau.

\end{document}